%% file: main.tex
\documentclass[11pt]{article}
\usepackage[letterpaper,margin=1in]{geometry}
\usepackage[T1]{fontenc}
\usepackage{lmodern}
\usepackage{latexsym}
\usepackage{url}
\usepackage{xurl}          %
\usepackage[round]{natbib}
\usepackage{booktabs}
\usepackage{adjustbox}     %
\usepackage{graphicx}
\usepackage{amsmath,amssymb,amsthm}
\usepackage{tikz}
\usepackage{microtype}
\usepackage{xcolor}
\usepackage[colorlinks=true,linkcolor=blue!50!black,citecolor=blue!50!black,
            urlcolor=blue!50!black]{hyperref}
\graphicspath{{figures/}}
\newtheorem{proposition}{Proposition}
\newtheorem{corollary}{Corollary}
\input{macros}                  %

\title{How Far Does Lossless Re-tokenization\\
       Move an RLVR Update?}
\author{Zhe Dong\thanks{Corresponding author.}\\
  University of Maine at Presque Isle
  \and
  Fang Qin\\
  Stanford University}
\date{}

\begin{document}
\maketitle

\begin{abstract}
Language models assign different probabilities to token sequences that
decode to the same string. In reinforcement learning from verifiable rewards
(RLVR), a trainer that re-encodes sampled text may train on a different
token sequence with the same reward. We ask how far this moves
a training update, and whether the movement has a consistent direction. From
one checkpoint, we compare updates on the same rollouts, canonical and with
selected tokens split into two pieces, against an update on a different batch.
On Qwen2.5-0.5B with three-digit addition, splitting about one token in
\IxOneTokenOneIn{} moves the update \IxOneTokenRandom{} times
as far as replacing the batch. The ratio rises with the fraction split in four
models and depends on which tokens are split; within a checkpoint, the
displacements share a direction across batches. In a tabular model with one
table per tokenization and a single sampled string, we prove that whenever two
strings below an unsampled prefix differ in probability, only the zero
full-gradient step agrees across tokenizations, however it is weighted. In an
audit that decodes and re-encodes resampled rollouts, re-encoding touches up
to \IxSidTopPctHi{} of rollouts but at most \IxSidMaxTokenPct{} of tokens, far
below the split fractions we test.
Training on fixed rollouts, the canonical arm finishes ahead in
\IxConfAhead{} of \IxConfSigned{} pre-registered pairs; the interval for
the mean accuracy difference includes zero over all pairs and excludes it
only under the registered exclusion rule, which favours the canonical arm.
\end{abstract}

\section{Introduction}

\begin{figure*}[t]
  \centering
  \includegraphics[width=\textwidth]{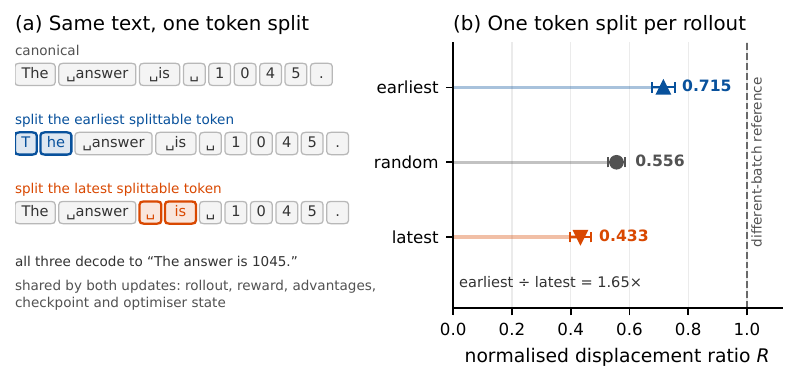}
  \caption{Splitting one token per rollout without changing the text.
  (a) A completion under the Qwen2.5 tokenizer, canonical and with its
  earliest or its latest splittable token split; \textvisiblespace{} marks a
  space. All three decode to the same string. (b) The ratio $R$
  (Equation~\ref{eq:R}): how far the split moves the update, relative to how
  far an update on a different batch from the same task moves it. Means over five seeds with $\pm1$ standard error. The policies
  split different tokens, so token identity as well as position can
  contribute. Qwen2.5-0.5B, three-digit addition, LoRA,
  \texttt{seq\_reinforce}, batches of \IxProbeBatch{} rollouts.}
  \label{fig:teaser}
\end{figure*}

A subword vocabulary usually offers more than one token sequence for the same
string. In the Qwen2.5 vocabulary, the word \emph{The} is a single token, but
its pieces \emph{T} and \emph{he} are tokens too, and both sequences decode to
the same text (Figure~\ref{fig:teaser}a). A language model assigns such
sequences different probabilities \citep{cao2021marginal,
chirkova2023marginalize, geh2024signal}. In reinforcement learning from
verifiable rewards (RLVR) \citep{lambert2024tulu3, guo2025deepseekr1}, this
ambiguity can reach the training update. Nothing stops the policy from
sampling a sequence the encoder would never produce, such as \emph{T}
followed by \emph{he}. If the trainer receives only the decoded text and
encodes it again, it computes the loss on \emph{The}, a token the policy did
not sample, although the text, and with it the reward, is unchanged.

Systems work counts this mismatch as one source of \emph{retokenization
drift} and avoids it, by carrying the sampled token IDs through the pipeline
\citep{vllm2025agentlightning, hf2026tito, lmsys2026notokenleftbehind} or by
encoding the segments of a rollout separately \citep{verltool}. These studies
give good reasons to preserve token IDs but do not measure what a mismatch
does to an update. With the text, reward and advantage of every rollout held
fixed, two questions decide whether a change of tokenization matters. How far
does it move the update, compared with the ordinary variation between
batches? And does it move the update in the same direction from batch to
batch, so that its effect could add up over training instead of averaging
out?

We answer both with a paired measurement of a single update
(Section~\ref{sec:measurement}). We call any change of tokenization that keeps
the text a \emph{re-encoding}. Where the example above merges \emph{T} and
\emph{he}, our intervention splits selected tokens into two vocabulary items
with the same text (Figure~\ref{fig:teaser}a), so that we control how many
tokens change and which. From one checkpoint, we compute one update on a
batch of rollouts in its canonical encoding and one on its split version. We
compare their distance with the distance between the canonical update and an
update on a different batch from the same task, each divided by the average
norm of the two updates it compares. The ratio $R$ of the two distances is one
when re-encoding a batch moves the update as far as replacing the batch
does. For the second
question, we measure whether the differences between the two updates point
the same way across batches.

On Qwen2.5-0.5B learning three-digit addition, splitting one random splittable
token per rollout, about one token in \IxOneTokenOneIn{}, gives
$R = \IxOneTokenRandom{}$: the update moves about half as far as on a
different batch (Figure~\ref{fig:teaser}b). Splitting each rollout's
earliest splittable token instead moves it \IxOneTokenFrontBackRatio{} times
as far as splitting its latest. $R$ rises with the fraction of tokens split in
all four models we test on addition, reaching about one in the Qwen models
when a third of the tokens are split, and it rises on GSM8K as well.
Full fine-tuning lowers $R$ but leaves it substantial, and neither the bare
gradient nor any of five surrogate objectives removes the effect
(Section~\ref{sec:dose-location}).

The differences also share a direction. Their alignment across batches, a
normalised mean dot product that is zero for unrelated directions, reaches
\IxAlignLoraHigh{} when a third of the tokens are split and \IxOneAlignFront{}
when only the earliest splittable token of each rollout is, against \IxAlignFloorMean{}, pooled over all cells, for the differences
caused by replacing the batch, and it stays above that level at later
checkpoints (Section~\ref{sec:align}). Re-encoding therefore does more than
add batch-like variation.

In a tabular model, reweighting cannot in general remove the effect
(Section~\ref{sec:theory}). Take two tokenizations of the same strings, both
starting from the same distribution over strings. With one softmax row over
whole strings, a gradient step on a sampled string changes the ratios between
unsampled strings, because each unsampled logit moves in proportion to its
probability. With one row per prefix, the step changes only the rows it
visits, never the row below an unsampled prefix, so the ratios read from that
row stay fixed, and no weight can change that. Whenever two strings below
such a prefix differ in probability, the two tokenizations therefore agree
only on the zero step, whether the step is weighted per token, by length, by
a sequence-level importance ratio or by clipping.

How much re-encoding does a real pipeline produce? When we decode and
re-encode rollouts resampled from all four models, as a trainer receiving only
text would, up to \IxSidTopPctHi{} of rollouts change, but in every setting
at most \IxSidMaxTokenPct{} of all sampled tokens, far below our smallest
split fraction, \IxSidSmallestDose{} (Section~\ref{sec:audit}). Our results
therefore describe how sensitive an update is to re-encoding, not what
re-encoding costs a pipeline. We also test whether this sensitivity changes
the outcome of training. In a pre-registered comparison on GSM8K, two arms
train on one fixed pool of rollouts, one on the sampled tokens and one with a
tenth of them split. The arm trained on the sampled tokens finishes ahead in
\IxConfAhead{} of \IxConfSigned{} pairs, but the $95\%$ interval for the mean
accuracy difference includes zero. The registered analysis, which drops
\IxConfExDropped{} pairs by the canonical arm's accuracy, excludes zero, but
its exclusion rule favours the canonical arm (Section~\ref{sec:outcome}).

\paragraph{Contributions.} (i) A paired single-update measurement, with a
different-batch reference, of how far lossless re-encoding moves an RLVR
update (Section~\ref{sec:measurement}). (ii) Its dependence on dose and edit
location, its persistence across parameter spaces, update rules and
objectives, and its shared direction across batches
(Section~\ref{sec:dose-location}). (iii) A proof that, in a tabular model
with one table per tokenization, no weighting of a full-gradient step makes
the tokenizations agree except the zero step, wherever the strings below an
unsampled prefix differ in probability (Section~\ref{sec:theory}). (iv) An
audit of sampled token IDs that separates how many rollouts a re-encoding
touches from how many tokens it edits (Section~\ref{sec:audit}).

\section{Measuring the effect of a re-encoding}
\label{sec:measurement}
\label{sec:setup}

\subsection{Two encodings of one string}
A BPE vocabulary \citep{sennrich2016bpe} builds long tokens by merging shorter
ones, and the merged pieces stay in the vocabulary. A token of more than one
character can therefore often be cut into two vocabulary items whose texts
join to form its own text. Replacing the token by the two items gives a second
encoding of the same string, undoing a merge where the introduction's example
makes one. We study subword vocabularies
that allow such splits; byte- and character-level models
\citep{xue2022byt5, clark2022canine} give each string a single encoding.

We call a token splittable if its text can be cut into two vocabulary items,
and use the leftmost valid cut, so one token always becomes exactly two
(Appendix~\ref{app:method}). The \emph{dose}, or split fraction, is the
fraction of a rollout's sampled tokens that are split. Requested doses are
rounded at random to whole splits per rollout, and we report the achieved
fraction, which falls short when a rollout has too few splittable tokens. By default the split tokens are drawn
at random among the splittable ones; the location experiments of
Section~\ref{sec:position} instead take the earliest or the latest splittable
tokens of each rollout, up to the requested number. The two encodings of every
rollout decode to exactly the same string, which we check with the tokenizer's
clean-up of spaces at decoding turned off.

\subsection{Paired updates}
\label{sec:measure}
We measure a single update from a shared checkpoint, so the two arms have no
trajectory along which to drift apart; Section~\ref{sec:outcome} shows how
noisy completed training runs are. Rollouts are sampled once from the initial
checkpoint into a pool, from which both the warmup and the measured batches
are drawn. The warmup is a few updates on canonical encodings, which lets both
LoRA factors receive gradients (Appendix~\ref{app:method}). From the resulting
checkpoint we apply one update to the canonical encodings of a batch and one
to its split encodings. The two arms share their prompts, rollouts, rewards
and advantages and differ only in the token sequences. The canonical arm
trains on the token IDs as sampled, as a pipeline that preserves them would;
in the first-seed probe pools, which we re-drew, these equal the encoder's
encoding of the text for all but \IxSidPoolAddChanged{} of \IxSidPoolAddN{}
addition rollouts and \IxSidPoolGsmChanged{} of \IxSidPoolGsmN{} GSM8K
rollouts (Appendix~\ref{app:audits}).

Each update consists of \IxInnerEpochs{} inner optimisation steps on the
same batch, using AdamW \citep{loshchilov2019adamw} with its state reset
before the first step and kept between steps. As in GRPO \citep{shao2024deepseekmath},
advantages are group-relative, and objectives with importance ratios take
them against the checkpoint, frozen as the old policy $\pi_{\mathrm{old}}$.
Both the current and the old policy are evaluated on the arm's own token
sequence, as in a trainer that re-encodes, so the ratios do not correct for
the change of tokenization. We write $\Delta W$ for the
resulting net change in the trainable parameters; under LoRA
\citep{hu2022lora}, these are the adapter factors, not the merged weight
matrices.

The update uses one of five surrogate objectives, which share GRPO's
group-relative advantages and differ in how they weight tokens and form
importance ratios (Appendix~\ref{app:method}). At the first inner step every
importance ratio equals one and no clipping is active, so the five objectives
reduce to two gradients, of the summed and of the length-averaged token
log-probabilities; the later inner steps are what separate them.
\texttt{token\_mean} is GRPO's clipped surrogate without its KL term.
\texttt{seq\_reinforce}, which we report unless stated otherwise, multiplies
each rollout's summed token log-probability by its advantage, with no
importance ratio, clipping or per-token weight. The tokenization enters it
only through which tokens are summed.

Because the two arms score different token sequences, we compare the
parameter updates they produce, by their distance divided by their average
length:
\begin{equation}
c \;=\; \frac{\lVert \Delta W_{\mathrm{canon}} - \Delta W_{\mathrm{re}} \rVert}
              {\tfrac12\left(\lVert \Delta W_{\mathrm{canon}}\rVert
                           + \lVert \Delta W_{\mathrm{re}}\rVert\right)} .
\label{eq:c}
\end{equation}
Scaling both updates by a common factor leaves $c$ unchanged, and in a sweep
over four learning rates the distance between the two updates grew in
proportion to their length (fitted exponent $\alpha = \IxStepAlpha{}$,
Figure~\ref{fig:steps}), so dividing by the length removes the step size. The two updates also have nearly the same length:
over \IxNNormRatio{} cells, where a cell is one seed at one setting, the norm
of the re-encoded update divided by that of the canonical one averages
\IxNormRatio{}. $c$ therefore mostly reflects a difference in direction, and
we also report the cosine similarity $\cos\theta$ between the canonical and
the re-encoded update. Section~\ref{sec:space} repeats the measurement
without the optimiser.

\subsection{A reference scale}
\label{sec:references}
A value of $c$ on its own does not say whether a displacement is large. Our
reference is the displacement caused by replacing the data: an update, from
the same checkpoint, on a disjoint batch of unmodified rollouts from the same
task. Replacing the batch changes the rollouts and usually the prompts too.
We report re-encoding as the ratio of mean distances
\begin{equation}
R \;=\; \frac{\operatorname{mean}_b\, c_{\mathrm{re},b}}{\operatorname{mean}_b\, c_{\mathrm{ref},b}},
\label{eq:R}
\end{equation}
where $c_{\mathrm{re},b}$ and $c_{\mathrm{ref},b}$ are Equation~(\ref{eq:c})
between the canonical update on batch $b$ and, respectively, the re-encoded
update and the update on the disjoint batch, and the means run over the
batches where both were measured. $R$ is computed within each cell and then
averaged across seeds. $R = 1$ means that re-encoding a batch displaces the
update as much as replacing it with different data does.

Two geometric landmarks help to read $c$. Repeating the same batch in the
same order with the same tokens gives $c = 0$, as it did exactly in all
\IxNSameOrderNull{} runs where we checked; two updates of equal length in
orthogonal directions give $c = \sqrt{2}$, and opposite updates give $2$. The
reference lies well inside this range: averaged over the Qwen2.5-0.5B
addition cells under LoRA, its $c$ is \IxNullValueLoRA{} and the cosine between the canonical update and the
update on the disjoint batch is \IxNullCosLoRA{}. $R$ near one therefore marks
a displacement as large as that of a different batch, which is itself large,
not one close to the largest possible $c$. We estimate $R$ at a batch size of \IxProbeBatch{} rollouts;
Section~\ref{sec:discussion} discusses larger batches.

\section{Sensitivity of a single update}
\label{sec:dose-location}
\label{sec:size}
\label{sec:update-rule}

Figure~\ref{fig:teaser}b fixes the model, the task, the trainable parameters,
the objective and the number of split tokens. This section varies each of
them, and then asks whether the displacements share a direction across
batches.

\subsection{Dose}
\label{sec:dose}
\begin{figure*}[t]
  \centering
  \includegraphics[width=\linewidth]{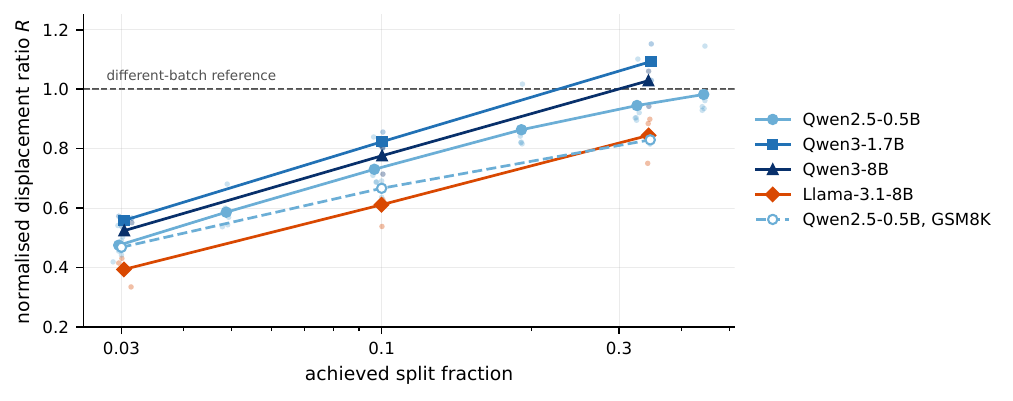}
  \caption{The ratio $R$ (Equation~\ref{eq:R}) against the achieved split
  fraction, for four models on three-digit addition and for Qwen2.5-0.5B on
  GSM8K (hollow markers, dashed line). Small points are seeds; large points are
  means over seeds at each requested dose, placed at the mean achieved
  fraction. The three Qwen models share one vocabulary and Llama-3.1-8B has
  another (sizes in Table~\ref{tab:scale}). LoRA, \texttt{seq\_reinforce};
  five seeds for Qwen2.5-0.5B on addition and three for every other curve.}
  \label{fig:dose}
\end{figure*}

$R$ grows with the dose in every model we test (Figure~\ref{fig:dose}). On
Qwen2.5-0.5B with addition, it rises from \IxNullRatioQwenHalfBLo{} at a split
fraction of \IxSidSmallestDose{} to \IxNullRatioQwenHalfBHi{} when about a
third of the tokens are split. At that dose the two Qwen3 models reach or
exceed the reference, whereas Llama-3.1-8B stays below it
\citep{qwen2024qwen25, qwen2025qwen3, grattafiori2024llama3};
Table~\ref{tab:scale} in Appendix~\ref{app:heldout} gives three doses for
each model. On GSM8K \citep{cobbe2021gsm8k}, $R$ rises
with the dose as well (dashed curve); averaged over its three doses it is
\IxNullRatioGSMEIGHTK{}, against \IxNullRatioLoRAAtGSMDoses{} on addition at
the same doses.

The three Qwen models share a vocabulary and span a factor of $16$ in
parameters, yet their means do not follow size: at every dose Qwen3-1.7B is
highest and Qwen2.5-0.5B lowest, and their seed ranges overlap, so these runs
detect no trend with model size. Llama-3.1-8B has the lowest mean at every
dose, and at the middle dose its seed range does not overlap those of the
Qwen models. With a single Llama model, however, its vocabulary and
everything else about it vary together, so the difference cannot be
attributed to the vocabulary.

As the dose increases, the canonical and the re-encoded update also point in
increasingly different directions. Over the tested range, their cosine
similarity follows a power law in the achieved split fraction $f$,
$\cos\theta \approx \kappa f^{\gamma}$ (Figure~\ref{fig:doselaw} in
Appendix~\ref{app:heldout}), with $\gamma$ between \IxDoseExpMax{} and
\IxDoseExpMin{} across the four models; the fit with the most doses,
Qwen2.5-0.5B on addition with \IxNDosesQwenHalfB{} doses, gives
\IxDoseExpQwenHalfB{} $\pm$ \IxDoseExpSEQwenHalfB{}. The two extreme exponents
differ by \IxDoseExpExtremeZ{} times their combined standard error, so we
report one per model.

\subsection{Location}
\label{sec:position}
\begin{figure*}[t]
  \centering
  \includegraphics[width=\linewidth]{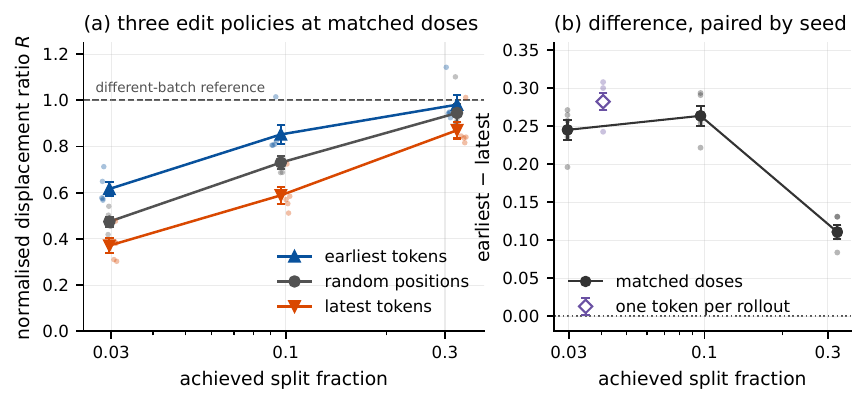}
  \caption{Edit policies at matched requested doses, on Qwen2.5-0.5B with addition
  (LoRA, \texttt{seq\_reinforce}, five seeds). (a) The ratio $R$ when the
  earliest, randomly placed or latest splittable tokens are split.
  Large points are means over seeds with $\pm1$ standard error, placed at the
  mean achieved split fraction of each requested dose; seed points are offset
  sideways by policy.
  (b) Earliest minus latest in $R$, computed within each seed and then
  summarised (mean $\pm1$ standard error), at the three doses and for one
  split token per rollout (hollow diamond).}
  \label{fig:position}
\end{figure*}

At each of the three requested doses, splitting the earliest splittable
tokens of each rollout moves the update furthest and splitting the latest
moves it least, with random positions in between; at the lowest dose the
three give \IxPosFirstLo{}, \IxPosRandomLo{} and \IxPosLastLo{}
(Figure~\ref{fig:position}a). The earliest and the latest splittable tokens
can differ in identity and in probability under the policy, so this compares
edit policies rather than isolating position. Paired by seed, earliest minus
latest is \IxPosDiffLo{}, \IxPosDiffMid{} and \IxPosDiffHi{} at the three
doses, positive in all \IxPosDiffN{} seeds at each
(Figure~\ref{fig:position}b): the gap narrows when a third of the tokens are
split, where all three policies approach the reference. With exactly one
token split per rollout, every policy has the same split fraction,
\IxOneTokenSplit{}, and the gap remains (Figure~\ref{fig:teaser}b). To match
the earliest-token value by splitting random tokens instead, interpolation
along the random-position curve gives a split fraction of \IxPosDoseEquiv{},
\IxPosDoseFold{} times as many tokens.

One natural explanation is that an early split matters more because more of
the rollout follows it and is conditioned on a changed context. If that were
the whole story, the three policies would give similar $R$ at the same
fraction of tokens after the first split. They do not: plotted against that
downstream fraction, they fall on different curves, and in regressions over
the same cells the split fraction fits $R$ better than the downstream fraction
does.

\subsection{Parameter space and optimiser}
\label{sec:space}
Training every parameter instead of a LoRA adapter lowers $R$ but does not
remove the effect. On Qwen2.5-0.5B with addition, averaged over the same
\IxNDosesQwenHalfB{} doses and five seeds, $R$ is \IxNullRatioFullFT{} under
full fine-tuning against \IxNullRatioLoRA{} under LoRA, and it is lower at
every dose. The size of the effect thus depends on which parameters are
trained.

The effect also persists without the optimiser. AdamW's first step after a
reset moves each coordinate by about $\mathrm{lr} \cdot \mathrm{sign}(g)$,
whatever the size of the gradient $g$, so it tends to equalise the lengths of
the two updates and can exaggerate a small edit: a coordinate whose gradient
the edit pushes across zero moves a full step the other way. Replacing each
update by the loss gradient at the shared checkpoint removes the optimiser,
together with the inner steps; we did so in six of the conditions above, five
seeds each (Table~\ref{tab:space}). At the lowest dose, $R$ on the gradient is lower in
every seed, \IxSpaceDoseLoGradient{} against \IxSpaceDoseLoUpdate{}; at a third
of the tokens it is slightly higher, \IxSpaceDoseHiGradient{} against
\IxSpaceDoseHiUpdate{}.

The contrasts of the previous two subsections are larger on the gradient, not
smaller. $R$ rises \IxSpaceDoseFoldGradient{}-fold from the lowest dose to the
highest, against \IxSpaceDoseFoldUpdate{}-fold for the optimiser step, and the
earliest-to-latest factor for one split token is \IxSpaceFrontBackGradient{}
against \IxSpaceFrontBackUpdate{}. Across all \IxSpaceN{} paired cells, the
ratios on the gradient and on the step are close to linearly related (slope
\IxSpaceSlope{}, $r^2 = \IxSpaceRTwo$). Relative to the gradient, the
four-step AdamW update thus compresses these contrasts rather than creating
them. We report the optimiser step elsewhere because it is what a trainer
applies. This check covers one model, task, adapter and objective.

\begin{table}[t]
  \centering
  \caption{The ratio $R$ with the AdamW update replaced by the loss gradient,
  paired by seed and condition (Qwen2.5-0.5B, addition, LoRA,
  \texttt{seq\_reinforce}). Each column is measured against the
  different-batch reference in its own space. \emph{shift} is gradient minus update, \emph{ratio} is gradient over
  update, and \emph{lower} counts the seeds in which the gradient value is the
  smaller.}
  \adjustbox{max width=\columnwidth}{\input{tables/space}}
  \label{tab:space}
\end{table}

\subsection{Objective}
\label{sec:objective}
Table~\ref{tab:objective} compares the canonical and the re-encoded update
under each objective by $c$, which normalises away the objectives' different
step lengths (Section~\ref{sec:measure}): under LoRA the
\texttt{seq\_reinforce} update is \IxStepRatioLoRA{} times as long as the
\texttt{token\_mean} update. \texttt{seq\_reinforce} has the smallest $c$
under both LoRA and full fine-tuning. \texttt{seq\_geom}, which is also
sequence-level but weights each rollout by the geometric mean of its token
importance ratios, gives a larger $c$ than \texttt{seq\_reinforce} in both
settings, so being sequence-level does not by itself produce the reduction.
Under full fine-tuning $c$ is about a fifth lower for \texttt{seq\_reinforce}
than for \texttt{token\_mean}, and under LoRA less than a tenth lower. The two
objectives also differ in importance ratios, clipping and length
normalisation, so the comparison does not identify which difference is
responsible. Every objective leaves a substantial displacement.

\begin{table}[t]
  \centering
  \caption{The normalised displacement $c$ (Equation~\ref{eq:c}) under each
  objective, ordered by its value under LoRA, over the dose sweep of
  Section~\ref{sec:dose} (Qwen2.5-0.5B on addition, \IxNDosesQwenHalfB{}
  doses, five seeds). \emph{step ratio} is the mean update length under
  \texttt{seq\_reinforce} over that under \texttt{token\_mean}.}
  \adjustbox{max width=\columnwidth}{\input{tables/objective}}
  \label{tab:objective}
\end{table}

\subsection{Alignment across batches}
\label{sec:align}
A displacement that is large compared with the reference could still be
noise that cancels when many batches are combined. Let $d_i =
\Delta W_{\mathrm{canon},i} - \Delta W_{\mathrm{re},i}$ be the re-encoding
displacement on batch $i$, with all updates starting from the same
checkpoint. Then
\begin{equation}
\Bigl\lVert \sum_i d_i \Bigr\rVert^2 \;=\; \sum_i \lVert d_i \rVert^2
  \;+\; \sum_{i \neq j} \langle d_i, d_j \rangle .
\end{equation}
For displacements with no shared direction the cross terms are zero on
average and the sum grows only through the first term; positive cross terms
make the displacements reinforce one another. We summarise them by the
alignment statistic $\rho$: the mean of $\langle d_i, d_j \rangle$ over
distinct pairs of batches, divided by the mean of $\lVert d_i \rVert^2$ over
batches. When all displacements have the same norm, $\rho$ is their mean
pairwise cosine, $0$ for no shared direction on average and $1$ for identical
displacements. As a control, we compute the same statistic for the canonical
minus the different-batch update, $\Delta W_{\mathrm{canon}} -
\Delta W_{\mathrm{ref}}$, which involves no re-encoding.

The re-encoding displacements share a direction, increasingly so as the dose
grows (Table~\ref{tab:align}). Under LoRA, $\rho$ rises from \IxAlignLoraLow{}
at a split fraction of \IxAlignLoraRiseLoDose{} to \IxAlignLoraHigh{} at
\IxAlignLoraRiseHiDose{}, in all \IxAlignLoraRiseN{} seeds (one-sided sign
test, $p = \IxAlignLoraRiseP$, the smallest value five seeds allow). Under full fine-tuning, measured at
\IxAlignFtRiseLoDose{} and \IxAlignFtRiseHiDose{} only, it rises less, from
\IxAlignFtMid{} to \IxAlignFtHigh{}, again in all \IxAlignFtRiseN{} seeds.
Pooled over all cells the control averages \IxAlignFloorMean{}, but its
standard deviation across cells, \IxAlignFloorSD{}, is as large as the
treatment at the lowest dose, so at that dose the evidence is the rise with
the dose rather than the value itself. With one split
token per rollout, $\rho$ is \IxOneAlignFront{} for the earliest splittable
token, \IxOneAlignFrontRatio{} times the control and above it in all
\IxOneAlignFrontN{} seeds, and \IxOneAlignBack{} for the latest.

The alignment persists at later checkpoints. Repeating the highest LoRA dose
after warmups of \IxWarmupLo{}, \IxWarmupMid{} and \IxWarmupHi{} canonical
updates gives $\rho$ of \IxWarmupAlignLo{}, \IxWarmupAlignMid{} and
\IxWarmupAlignHi{}, falling from the first checkpoint to the second in all
\IxWarmupDropNEightThirtyTwo{} seeds and changing little after that. At every
checkpoint $\rho$ exceeds the control in all \IxWarmupSeeds{} seeds, by a
factor of \IxWarmupRatioLo{} to \IxWarmupRatioHi{}. Each value describes
the batches at one checkpoint; whether the shared direction is the same at
different checkpoints was not measured, which matters for what the alignment
implies over a training run (Section~\ref{sec:discussion}).

\begin{table}[t]
  \centering
  \caption{The alignment statistic $\rho$ of the per-batch re-encoding
  displacements on Qwen2.5-0.5B with addition. The control is the same
  statistic for the canonical minus the different-batch update, which involves
  no re-encoding; it takes the same value at every dose of a seed.
  $\rho$ is computed within each cell from its \IxAlignBatchesLo{} to
  \IxAlignBatchesHi{} batches and then averaged over seeds. Full fine-tuning
  was measured at two doses only.}
  \adjustbox{max width=\columnwidth}{\input{tables/alignment}}
  \label{tab:align}
\end{table}

\section{Why reweighting cannot remove the dependence in a tabular model}
\label{sec:theory}
Every objective in Section~\ref{sec:objective} leaves a substantial
displacement. Could some other weighting of a gradient step make the updated
distribution over strings the same under both tokenizations? We study this in
a tabular model (Appendix~\ref{app:simulator}) in which each tokenization has
its own table of logits, one softmax row per token prefix. Both tables are
built from the same distribution over strings, so only the update can make
them differ; a pretrained neural model already gives two segmentations of a
string different probabilities \citep{geh2024signal}. Each
objective depends on the logits only through the log-probabilities of sampled
tokens, so a gradient step adds $\lambda_t\,(e_{z_t} - \pi)$ to the row of each
sampled token $z_t$, where $e_{z_t}$ is the token's one-hot vector and $\pi$ the
row's probabilities before the step. The scalar $\lambda_t$, which absorbs the
learning rate, is the weighting; it may depend on the token's probability,
the advantage, the length or a sequence-level ratio.

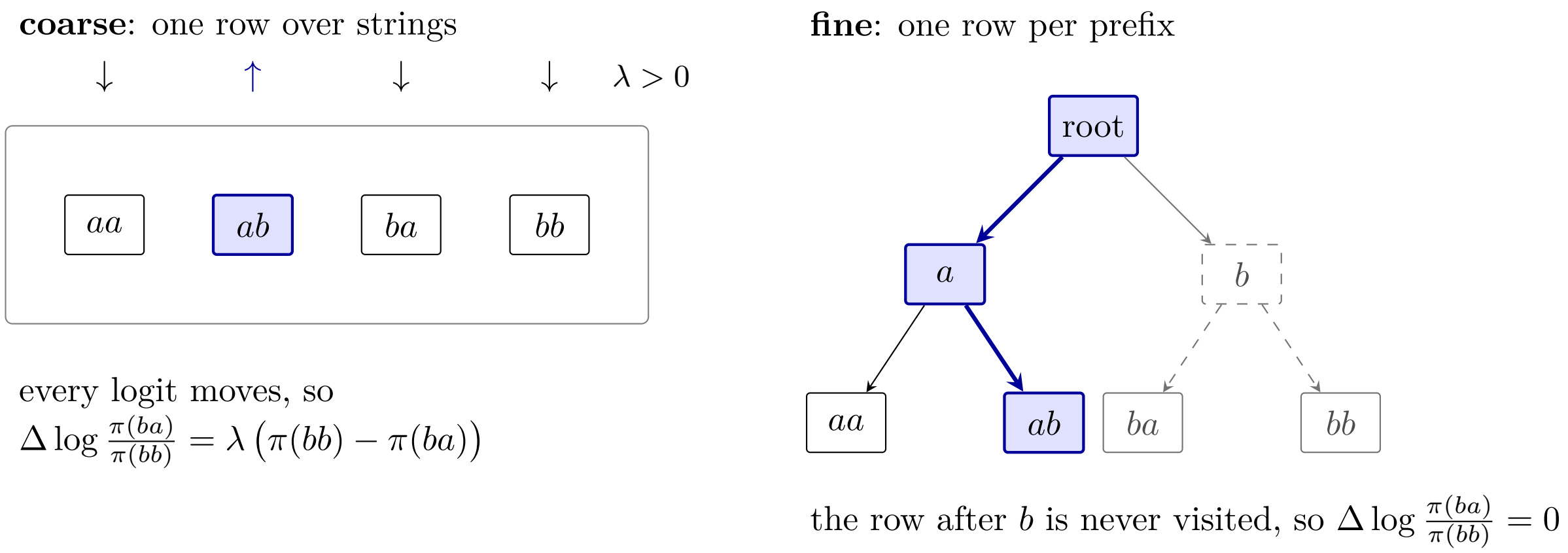
\begin{figure}[!htb]
  \centering
  \begin{tikzpicture}[x=5mm, y=5mm, font=\small, >=stealth,
      str/.style={draw, rounded corners=1pt, minimum width=8mm, minimum height=6mm},
      hit/.style={str, fill=blue!12, draw=blue!60!black, thick},
      row/.style={draw, rounded corners=1pt, minimum width=8mm, minimum height=6mm},
      path/.style={row, fill=blue!12, draw=blue!60!black, thick},
      idle/.style={row, draw=black!55, text=black!70, dashed}]
    \node[anchor=west] at (0,10) {\textbf{coarse}: one row over strings};
    \draw[rounded corners=2pt, black!50] (0,4) rectangle (13,8);
    \node[str] (caa) at (2,6) {$aa$};
    \node[hit] (cab) at (5,6) {$ab$};
    \node[str] (cba) at (8,6) {$ba$};
    \node[str] (cbb) at (11,6) {$bb$};
    \node[blue!60!black] at (5,9) {$\uparrow$};
    \foreach \n in {2,8,11} \node at (\n,9) {$\downarrow$};
    \node[anchor=west, font=\footnotesize] at (12,9) {$\lambda > 0$};
    \node[anchor=west, align=left] at (0,2)
      {every logit moves, so\\[1pt]
       $\Delta\log\frac{\pi(ba)}{\pi(bb)} = \lambda\,\bigl(\pi(bb)-\pi(ba)\bigr)$};
    \node[anchor=west] at (16,10) {\textbf{fine}: one row per prefix};
    \node[path] (r) at (22,8) {root};
    \node[path] (a) at (19,5) {$a$};
    \node[idle] (b) at (25,5) {$b$};
    \node[str] (faa) at (17,2) {$aa$};
    \node[hit] (fab) at (21,2) {$ab$};
    \node[str, draw=black!55, text=black!70] (fba) at (23,2) {$ba$};
    \node[str, draw=black!55, text=black!70] (fbb) at (27,2) {$bb$};
    \draw[->, very thick, blue!60!black] (r) -- (a);
    \draw[->, black!55] (r) -- (b);
    \draw[->] (a) -- (faa);
    \draw[->, very thick, blue!60!black] (a) -- (fab);
    \draw[->, black!55, dashed] (b) -- (fba);
    \draw[->, black!55, dashed] (b) -- (fbb);
    \node[anchor=west, align=left] at (16,0)
      {the row after $b$ is never visited, so
       $\Delta\log\frac{\pi(ba)}{\pi(bb)} = 0$};
  \end{tikzpicture}
  \caption{The two tokenizations under a gradient step on the sampled
  string $ab$, drawn for $\lambda > 0$. Left: the coarse tokenization has one softmax
  row over the four strings, and the step moves every logit in it, so the
  ratio $\pi(ba)/\pi(bb)$ changes unless $\pi(ba) = \pi(bb)$. Right: the fine
  tokenization has one row per prefix, and the step changes only the rows on
  the sampled path (shaded), never the row after $b$, so that ratio stays
  fixed whatever the weighting.}
  \label{fig:mechanism}
\end{figure}

The smallest case shows the mechanism, and one sampled string is enough: a
weighting that worked for every batch would have to work for this one. Take
strings of two characters over $\{a, b\}$, one token per string in the coarse
tokenization and one per character in the fine one, and a sampled string $ab$
(Figure~\ref{fig:mechanism}). The coarse policy is a single row over the four
strings, and a step with scalar $\lambda$ changes the logit of each unsampled
string $s$ by $-\lambda\,\pi(s)$. Strings of different probability move by
different amounts, so the ratio of two unsampled strings changes:
\begin{equation}
\Delta \log \frac{\pi(ba)}{\pi(bb)} \;=\; \lambda\,\bigl(\pi(bb) - \pi(ba)\bigr).
\label{eq:cex}
\end{equation}
The fine policy has a root row, a row after $a$ and a row after $b$. Its step
changes only the rows on the sampled path, the root row by
$\lambda_{\mathrm{root}}\,(e_a - \pi_{\mathrm{root}})$ and the row after $a$ by
$\lambda_a\,(e_b - \pi_{\cdot \mid a})$. It never visits the row after $b$, from
which $\pi(ba)/\pi(bb) = \pi(a \mid b)/\pi(b \mid b)$ is read, so it leaves that
ratio exactly unchanged, whatever its scalars.

\begin{proposition}
\label{prop:cex}
In this model, let every string have positive probability and let $ab$ be
sampled. If $\pi(ba) \neq \pi(bb)$, a coarse step with scalar $\lambda$ and a
fine step with scalars $\lambda_{\mathrm{root}}$ and $\lambda_a$ give the same
updated string distribution only if $\lambda = \lambda_{\mathrm{root}} =
\lambda_a = 0$.
\end{proposition}

Agreement on $\pi(ba)/\pi(bb)$ forces $\lambda = 0$ by
Equation~(\ref{eq:cex}), so the coarse distribution does not move. The fine
step must then leave the distribution unchanged too. A nonzero
$\lambda_{\mathrm{root}}$ changes the first-character marginal $\pi(a)$, and a
nonzero $\lambda_a$ changes the conditional $\pi(b \mid a)$, so both scalars
must vanish (Appendix~\ref{app:simulator}). Nothing is assumed about how the
scalars are computed, so the proposition covers pointwise weights, length
normalisers, sequence-level ratios and clipping alike. Over several inner
steps on the same sampled string, each step contributes the same expression
evaluated at the current distribution, so to first order in the learning rate
the proposition applies to the sum of the scalars.

The condition $\pi(ba) \neq \pi(bb)$ cannot be dropped. When the two strings
below the unsampled prefix are equally likely, Equation~(\ref{eq:cex}) is zero
for every $\lambda$. Nonzero steps can then agree: from the uniform
distribution, the fine step with $\lambda_a = \lambda$ and
$\lambda_{\mathrm{root}} = \log\bigl((e^{\lambda} + 1)/2\bigr)$, chosen so that
the first-character marginal matches the coarse step's, reproduces the coarse
step exactly. A weighting rule, however, has to work at every starting
distribution, which the corollary makes precise.

\begin{corollary}
\label{cor:rule}
Let a weighting rule assign, for each initial distribution with positive
probabilities, scalars to the coarse and the fine step on the sampled string
$ab$, computed from any information. If the two updated string distributions
agree for every initial distribution, the scalars are zero wherever
$\pi(ba) \neq \pi(bb)$; if they also depend continuously on the initial
distribution, they are zero everywhere.
\end{corollary}

The second claim holds because the distributions with $\pi(ba) \neq \pi(bb)$
are dense, so a continuous rule that is zero on them is zero also where
$\pi(ba) = \pi(bb)$. Without continuity it need not be: a rule could take the
matching nonzero steps above wherever $\pi(ba) = \pi(bb)$ and zero steps
elsewhere.

The result needs the full gradient. If a step only raised each sampled
token's logit, dropping $-\pi$, the coarse step would leave the ratios between
unsampled strings alone, and a weighting that reads the rest of the sequence
could make the two tokenizations agree on this example. Even in that weaker
model, the one continuous weighting, up to scale, that changes the sampled string's
probability equally under both tokenizations leaves the rest of the
distribution further apart than any other weighting tested, in
\IxNShapesInvariantWorst{} of \IxNShapes{} refinement shapes
(Appendix~\ref{app:simulator}, Tables~\ref{tab:cauchy}
and~\ref{tab:simulator}).

The proposition concerns separate tables for the two tokenizations, not a
network whose shared parameters couple them; for a network, the evidence is
that all five objectives of Section~\ref{sec:objective} leave a substantial
displacement.

\section{How much lossless re-encoding pipelines produce}
\label{sec:audit}
The doses so far were chosen by us. To see what a pipeline that stores text
and re-encodes it would hand its trainer, we resampled \IxSidGroup{} rollouts
for each of \IxSidNPrompts{} prompts, with the sampler, settings and order of
random draws of the paired-update experiments. We did this for each of the
\IxSidNModels{} models in three settings: addition, and GSM8K with and without
the chat template. We then compared every sampled completion with the
encoder's encoding of its decoded text (Appendix~\ref{app:audits}).
Table~\ref{tab:sampledids} counts the completions whose token IDs differ,
leaving out a few whose difference is not one of tokenization: completions
cut inside a multi-byte character and completions that spell out a special
token in ordinary tokens.

\begin{table}[htb]
  \centering
  \caption{Resampled rollouts whose sampled token IDs differ from the encoder's
  encoding of their decoded text, out of \IxSidNRollouts{} per cell, with the
  percentage in parentheses. Completions cut inside a multi-byte character and
  completions that spell out a special token are excluded.}
  \input{tables/sampled_ids}
  \label{tab:sampledids}
\end{table}

Qwen2.5-0.5B on GSM8K is the exception: \IxSidTopPctLo{} and
\IxSidTopPctHi{} of its rollouts differ from their re-encoding, without and
with the chat template. In every other setting no more than \IxSidRestPct{}
do (Table~\ref{tab:sampledids}). Even where many rollouts differ, few tokens
do. Counting the sampled tokens inside differing spans, an affected rollout
differs in \IxSidEditsPerRollout{} tokens on average, so across all settings the differing tokens are at most a
fraction \IxSidMaxTokenFrac{} of those sampled, against a split fraction of
\IxSidSmallestDose{} at our smallest dose. Our dose sweep therefore measures
sensitivity at edit densities well above the rates in this audit.

The differences mostly run in the opposite direction to our intervention. A
trainer that re-encodes replaces the sampled IDs with the encoder's. Of the
\IxSidBlocks{} differing spans, each a maximal run of differing tokens with
possibly several per rollout, \IxSidBlocksMerge{} merge tokens that the
sampler emitted separately, \IxSidBlocksSame{} move a boundary without
changing the number of tokens, as when \texttt{H|AVING} becomes
\texttt{HAV|ING} \citep{vllm2025agentlightning}, and only \IxSidBlocksSplit{}
split. We could not test merging at a matched dose: requested merge doses of
\IxMergeReqLo{} to \IxMergeReqHi{} on real addition rollouts edited on average
only a fraction \IxMergeEdit{} of the tokens. In the batches that did contain a
merge, the displacement per edited token was of the same order as for
splitting; at this exposure that is a consistency check, not a measurement of
merging.

The two other sources of drift listed by \citet{vllm2025agentlightning} change
the text, not only its tokenization. In the cases we constructed
(Appendix~\ref{app:audits}), re-applying a chat template to a conversation
left the decoded text unchanged in only \IxDriftTemplateStringKept{} of
\IxDriftTemplateN{} cases, and re-serialising a tool call in
\IxDriftToolReorderStringKept{} of \IxDriftToolReorderN{}; in no case did the
tokens change while the text stayed the same. Tool boundaries rarely produce
lossless drift. In GSM8K's reference solutions, calculator annotations mark
where a tool writes into the text. Encoding the model's and the tool's
segments jointly rather than separately changes the tokens at
\IxBoundaryChanged{} of \IxBoundaryN{} boundaries
(\IxBoundaryTokenFracLo{}--\IxBoundaryTokenFracHi{} of tokens) under both
vocabularies tested; each time, joint encoding merged the `=' that ends the
model's segment with a leading `.' or `$-$' of the result.

\section{Training on fixed rollouts}
\label{sec:outcome}
\label{sec:align-outcome}
Does the sensitivity of a single update carry through to final accuracy? We
test this in one pre-registered design (Appendix~\ref{app:prereg}): GSM8K, a
split fraction of \IxConfDose{}, \texttt{token\_mean} and \IxConfN{} paired
seeds (Table~\ref{tab:configconfirm}). Rollouts are sampled once from the
initial checkpoint and both arms train on that fixed pool in the same order,
so the design tests training on fixed rollouts, not on-policy RLVR
(Appendix~\ref{app:method}). Each update freezes $\pi_{\mathrm{old}}$ from
the current weights, as in the probe, but takes a single optimiser step
instead of \IxInnerEpochs{}, so its importance ratios equal one where the gradient is taken.

\begin{figure*}[t]
  \centering
  \includegraphics[width=\linewidth]{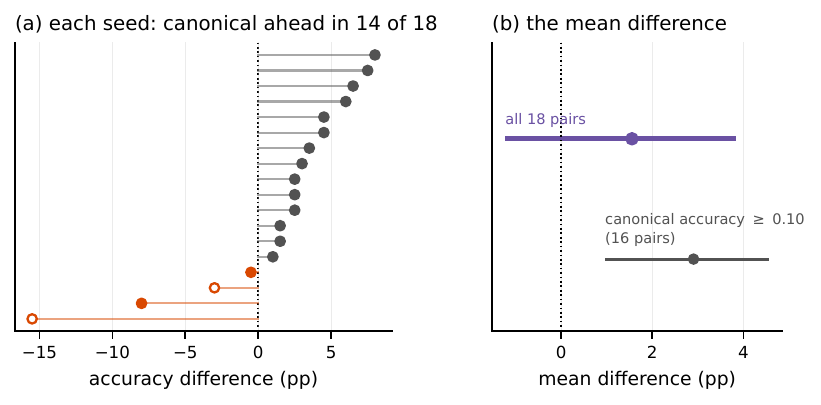}
  \caption{Final accuracy after training on a fixed pool of GSM8K rollouts,
  canonical minus re-encoded, in percentage points (pp); positive favours the
  canonical arm. (a) Each seed's paired difference, sorted; orange marks pairs
  in which the re-encoded arm finished ahead, and hollow points the two pairs
  the registered exclusion would drop. (b) The mean difference with a $95\%$
  paired bootstrap interval, over all \IxConfN{} pairs and over the pairs whose
  canonical arm reached \IxConfExThreshold{}.}
  \label{fig:outcome}
\end{figure*}

The canonical arm finishes with higher accuracy in \IxConfAhead{} of
\IxConfSigned{} pairs (two-sided sign test, $p = \IxConfSignP$). The mean
accuracy difference is \IxConfGapPP{} percentage points, with a $95\%$ paired
bootstrap interval from \IxConfLoPP{} to \IxConfHiPP{}
(Figure~\ref{fig:outcome}). The sign test thus supports a tendency for the
canonical arm to finish ahead more often, while the size of the mean
difference is not resolved and could be zero.

The registered primary analysis excluded pairs whose canonical arm finished
below \IxConfExThreshold{} accuracy. This drops \IxConfExDropped{} pairs and
gives a difference of \IxConfExGapPP{} points ($95\%$ interval \IxConfExLoPP{}
to \IxConfExHiPP{}). We lead with all pairs instead, departing from the
registration, because the rule selects on a component of the difference
itself: across the \IxConfN{} pairs, the canonical arm's final accuracy and the
paired difference correlate at \IxConfSelCorr{}, so dropping the pairs in which
that accuracy is low favours positive differences. Both excluded pairs favour
the re-encoded arm.

Final accuracy is also very sensitive to training randomness. In
\IxConfNullN{} control pairs on the same model and task, with both arms
trained on canonical encodings and differing only in minibatch order, the
paired final-accuracy differences have a standard deviation of
\IxConfNullSDPP{} percentage points.

Earlier training runs, which were not planned as a comparison, are
descriptive only. Of the \IxNTrainingRuns{} paired runs we completed,
\IxAugNRuns{} compare a canonical with a re-encoded arm and do not have both
arms at the accuracy floor or both at the ceiling; the rest are controls or
runs in which they do. The canonical arm, which trains on the sampled IDs,
finishes ahead in \IxAugCanonAhead{} of these, the re-encoded arm in
\IxAugRefinedAhead{}, and \IxAugTies{} tie (Table~\ref{tab:trainedout}). These
runs pool tasks, objectives, doses and model sizes, and include runs in which
one arm sits at a floor or a ceiling.

\section{Related work}
\label{sec:related}

\paragraph{Retokenization drift in RLVR systems.} The term comes from the team
behind Agent Lightning \citep{luo2025agentlightning}, whose write-up on the
vLLM blog lists three sources: non-unique tokenization, tool-call
re-serialisation and chat-template divergence \citep{vllm2025agentlightning}. \citet{hf2026tito}
describe the consequence as a gradient on tokens the policy never produced,
\citet{lmsys2026notokenleftbehind} formalise the remedy as
Token-In-Token-Out, and recent systems carry the sampled token IDs through to the trainer
\citep{prorlagent, polar}. \citet{verltool} locate the mismatch at individual
token IDs, where joint tokenization merges boundary symbols that the sampler
emitted separately, and prescribe tokenizing the segments apart. These works
prevent the mismatch; we measure how far a lossless change of token IDs moves
an update.

A different disagreement arises when the rollout engine and the trainer
assign different probabilities to the \emph{same} token sequence, because
their kernels and numerical precision differ. The remedy there is a
truncated importance-sampling correction by the ratio of the trainer's to the engine's
probability of each token \citep{yao2025rollouttraining}. Under re-encoding
the two sides do not share a token sequence, so that correction does not
apply directly, and Section~\ref{sec:theory} shows that in a tabular model
reweighting a gradient step cannot make two tokenizations agree except with a
zero step.

\paragraph{Marginalising over tokenizations.} Marginalising over
tokenizations improves likelihood \citep{cao2021marginal}, though in most
cases by under half a percent in log-likelihood over scoring the canonical
tokenization alone \citep{chirkova2023marginalize}. \citet{geh2024signal}
prove that finding the most likely tokenization and marginalising over all of
them are computationally hard, and show that the alternatives carry signal
that helps when aggregated. A model trained only on canonical encodings still
gives the others probability \citep{vieira2025canonical}, so a sampler can
emit them. Tokenization also biases prediction by itself: \citet{phan2025exact}
show that a tokenized model and a byte-level model that are statistically
equivalent can still predict the next byte substantially differently.

\paragraph{Robustness to non-canonical tokenizations.} Models are largely
robust to non-canonical tokenizations at inference, less so for those further
from canonical \citep{zheng2025brokentokens}, and this robustness partly
emerges during training \citep{jain2026retokenization}. Behaviour taught in
post-training need not carry over, however: a non-canonical tokenization of an
unsafe request can elicit a response that the model refuses under the
canonical one \citep{geh2025adversarial}. Robustness at inference does not by
itself imply that training updates are invariant, which is what we measure.

\paragraph{Sampling segmentations during training.} Subword regularization
samples among segmentations during training \citep{kudo2018subword}, and
BPE-dropout perturbs the merge procedure \citep{provilkov2020bpedropout}. Both
report gains. There, the segmentation is chosen before an example is scored,
and the loss is computed on it. Re-encoding in RLVR instead changes which
token IDs are scored after sampling, while the reward and advantage remain
those of the sampled rollout.

\paragraph{Impossibility results for weightings.} \citet{lengthinvariance}
prove that under outcome rewards no length-based weighting is both
gradient-unbiased and length-invariant. Both results limit what a weighting
can achieve, but theirs is a theorem about the policy-gradient estimator and
ours a result about reweighting a gradient step in a tabular model.

\section{Discussion and limitations}
\label{sec:discussion}
\label{sec:limitations}

\paragraph{What the ratio measures.} $R$ compares two displacements of one
update in parameter space. That splitting about one token in \IxOneTokenOneIn{} moves the
update about half as far as replacing the batch does says nothing directly
about learning or accuracy,
and the reference is itself a large perturbation: the canonical update and
the different-batch update have a cosine of only \IxNullCosLoRA{}
(Section~\ref{sec:references}). The evidence on accuracy comes from
Section~\ref{sec:outcome}, where the interval for the mean difference
includes zero over all pairs and excludes it only under the registered
exclusion rule.

\paragraph{Dose and exposure.} The split fraction is an intervention, not an
estimate of how often re-encoding happens. In our decode-and-re-encode audit
(Section~\ref{sec:audit}), lossless re-encoding touches up to
\IxSidTopPctHi{} of rollouts but far fewer tokens than any dose tested here,
and it mostly merges where our intervention splits. Template and tool-call
mismatches fall at segment boundaries, and Section~\ref{sec:position} shows
that where an edit falls matters, but because they change the text as well
as the tokens, our measurements do not quantify their effect.

\paragraph{From one update to many.} With $\rho$ as defined in
Section~\ref{sec:align} and $\overline{\lVert d\rVert^2}$ the mean squared
norm of the displacements, the sum of $N$ displacements has squared length
$N\,\overline{\lVert d\rVert^2}\,\bigl(1 + (N-1)\rho\bigr)$, so the cross
terms outweigh the sum of squared norms once $N > 1 + 1/\rho$. Under LoRA
this crossover lies near \IxAlignCrossLoraHigh{} batches at the highest dose
and \IxAlignCrossLoraLow{} at the lowest, near \IxAlignCrossFtHigh{} with every
parameter trained at the highest dose, and near \IxOneAlignFrontCross{} for
one split token at the front. Carrying
this from batches at one checkpoint to the steps of a training run would
require displacements of comparable size and a shared direction that persists
as training moves through checkpoints, whereas Section~\ref{sec:align}
measured alignment only within checkpoints.

\paragraph{Batch size.} The reference is the displacement from replacing one
batch of \IxProbeBatch{} rollouts with another, and training batches are
usually larger. For the loss gradient at one checkpoint, a batch of $kB$
rollouts averages $k$ batches of $B$, so the batch-to-batch noise shrinks as
$1/\sqrt{k}$ while the shared part of the re-encoding displacement does not.
Under the assumptions of Appendix~\ref{app:method}, this gives $R(kB)/R(B) =
\sqrt{1 + (k-1)\rho}$. Taking the alignment measured on the AdamW update as
an illustration, $\rho = \IxOneAlignRandom{}$ for one randomly placed split
token would double $R$ at about \IxBatchDoubleRandom{} rollouts per batch, and
$\rho = \IxOneAlignFront{}$ for one at the front at about
\IxBatchDoubleFront{}. If the prediction holds, our values understate $R$ at
training batch sizes. It remains a prediction for a batch-size sweep, since
the AdamW update is not linear in the batch and $\rho$ was measured at
$B = \IxProbeBatch{}$.

\paragraph{Other limits.} The outcome experiment covers one dose, task and
objective on fixed rollouts; the family comparison rests on one model outside
the Qwen family; and the reweighting result holds for separate tables, not for
a network whose shared parameters couple them. The achievable dose is also
capped by how many splittable tokens the policy writes, and a competent policy
can saturate a task, so that all rollouts of a prompt receive the same reward
and carry no advantage (Appendix~\ref{app:operating}).

\paragraph{What this establishes.} A paired single-update probe with a
different-batch reference turns the effect of a tokenization choice on an RLVR
update into numbers, the normalised displacement $c$ and its ratio $R$ to the
reference, that can be compared across models, objectives and training
setups. The update is measurably sensitive to lossless re-encoding: the
displacement rises with the dose in every model tested, depends on which
tokens are edited, persists under every objective and on the bare gradient,
and shares a direction across batches within a checkpoint. The probe needs no
completed training run; whether its rankings of tokenizers, pipelines or
objectives predict training outcomes is not yet tested.

\bibliography{refs}
\bibliographystyle{plainnat}

\appendix

\section{Pre-registration}
\label{app:prereg}
For each run, we recorded the claim under test, the primary and secondary
metrics and the CONTINUE/PIVOT/STOP thresholds in a gate file before inspecting
its results; each gate file is stored alongside the cell it judged.

The confirmatory run of Section~\ref{sec:outcome} has a full pre-registration,
written before the run and released with the code. It fixed the following.
\begin{itemize}
\item \emph{Claim.} Training on the re-encoded tokenization of the same
  rollouts yields lower final accuracy than training on the canonical one, on
  GSM8K at a split fraction of \IxConfDose{} under \texttt{token\_mean}.
\item \emph{Design.} Qwen2.5-0.5B-Instruct, LoRA rank 16, random split
  positions; the two arms of a pair share the seed, rollouts, rewards,
  initialisation and step count. One dose, so as not to reopen the
  multiplicity of the earlier corpus.
\item \emph{Sample size.} \IxConfN{} pairs: $80\%$ power at the $5\%$
  two-sided level for a gap of five points, from the paired-gap spread of the
  28 earlier GSM8K runs.
\item \emph{Exclusion.} Drop a pair if its canonical arm finishes below
  \IxConfExThreshold{} accuracy, and nothing else; below 14 surviving pairs the
  run is underpowered and no verdict is drawn.
\item \emph{Analysis.} Primary: the mean paired gap with a $95\%$ paired
  bootstrap interval, 10{,}000 resamples. Secondary: an exact two-sided sign
  test. Both analyses are reported regardless of their outcomes.
\item \emph{Decision.} An interval excluding zero on the positive side means
  re-encoding costs accuracy; on the negative side, that it helps; an interval
  including zero, that no effect is resolvable at this sample size.
\end{itemize}
Section~\ref{sec:outcome} departs from this in one respect: its headline
applies the primary analysis to every pair instead of after the exclusion,
because the exclusion conditions on the canonical arm. As registered, with the
exclusion, the interval excludes zero on the positive side, so the registered
verdict is that re-encoding costs accuracy; the section reports that estimate
beside the headline.

\section{Choosing the operating point}
\label{app:operating}
The probe requires variation in rewards within a prompt group. If all rollouts
in a group receive the same reward, their group-relative advantages are zero,
so the group contributes no update signal; this can occur when all rollouts are
incorrect or all are correct.

We screened model $\times$ task $\times$ prompt-format combinations for
nonzero within-group advantages, recording the fraction of rollouts carrying
one. Two facts from it bound the experiments. Digits are single-character
tokens with nothing to split, so applying a chat template to Qwen2.5-0.5B on
the multiplication task \texttt{mul3x1} drops the largest reachable split fraction from
\IxSplitCeilBare{} to \IxSplitCeilChat{}. The same competence saturates the
task: all rollouts in a group received the same reward in \IxNSaturated{} of
\IxNSaturatedTotal{} model-task combinations we checked, leaving no update
signal. That is why the competent-policy setting here is GSM8K rather than a
synthetic task.

\section{Configuration}
\label{app:config}
Table~\ref{tab:config} is generated directly from the per-cell records, like
every other number here. Where a setting varies across cells, every value it took is
listed. The paper names the checkpoints without their organisation and
\texttt{-Instruct} suffixes; Table~\ref{tab:config} gives the full IDs.
Table~\ref{tab:configconfirm} gives the confirmatory training runs, which
Table~\ref{tab:config} does not cover.

The addition prompt is \emph{Compute $a$ + $b$. Reply with only: The answer is
N}, and a rollout earns reward $1$ if it contains ``answer is'' followed by the
correct sum and no further digit. The GSM8K prompt is \emph{Solve the problem.
Reason briefly, then give the final number on its own line after `Answer:'},
followed by the problem, and a rollout earns reward $1$ if the last number it
writes equals the reference answer after removing thousands separators and
trailing zeros. No chat template is applied in either case, and training and
evaluation problems come from disjoint splits.

\begin{table*}[t]
  \centering
  \caption{Configuration of the paired probes. Rank, clip and top-$p$
  are read out of the training source; the rest are read from the probe cells,
  which is why the seed list is theirs and not the \IxConfN{} paired seeds of
  the confirmatory run or the counts in Table~\ref{tab:scale}.}
  \input{tables/config}
  \label{tab:config}
\end{table*}

\begin{table*}[t]
  \centering
  \caption{Configuration of the pre-registered training runs, read from the
  runs' records and the training source.}
  \input{tables/config_confirm}
  \label{tab:configconfirm}
\end{table*}

\section{Objectives and the probe}
\label{app:method}
\paragraph{Objectives.} A rollout $i$ has prompt $x_i$, group-normalised
advantage $A_i$ and an encoding $z^i = (z^i_1, \dots, z^i_{L_i})$: the
canonical one in one arm, the re-encoded one in the other. An update starts by
freezing $\pi_{\mathrm{old}}$, the current weights, for its \IxInnerEpochs{}
inner steps, and each arm evaluates token probabilities under this frozen
old policy on its \emph{own} encoding, as a trainer that re-encodes would.
$\pi_{\mathrm{old}}$ is the policy at the start of the update, not the policy
that sampled the rollouts. With
$r^i_t = \pi_\theta(z^i_t \mid x_i, z^i_{<t}) / \pi_{\mathrm{old}}(z^i_t \mid x_i, z^i_{<t})$,
$\phi(r, A) = \min\bigl(rA, \mathrm{clip}(r, 1 - \epsilon, 1 + \epsilon)A\bigr)$
as in PPO \citep{schulman2017ppo},
$\epsilon$ as in Table~\ref{tab:config} and $B$ rollouts per batch, the
objectives maximised are
\begin{align*}
\texttt{token\_sum} &: \tfrac1B \textstyle\sum_i \sum_t \phi(r^i_t, A_i), \\
\texttt{token\_mean} &: \tfrac1B \textstyle\sum_i \tfrac{1}{L_i} \sum_t \phi(r^i_t, A_i), \\
\texttt{seq}, \texttt{seq\_geom} &: \tfrac1B \textstyle\sum_i \phi(R_i, A_i), \\
\texttt{seq\_reinforce} &: \tfrac1B \textstyle\sum_i A_i \log \pi_\theta(z^i \mid x_i),
\end{align*}
where $\log R_i$ is $\sum_t \log r^i_t$ clamped to $[-10, 10]$ for
\texttt{seq} and $\frac{1}{L_i}\sum_t \log r^i_t$ for \texttt{seq\_geom}, the
length-normalised sequence ratio of GSPO \citep{zheng2025gspo}.
The clamp keeps the exponent of \texttt{seq} finite on a real rollout. The last
objective has no ratio and no per-token weight, and is left uncorrected across
inner steps, since a correction would put a ratio back.

\paragraph{Splitting a token.} A token is splittable if its string can be cut
into two vocabulary items that decode to the same text; we take the first such
cut, scanning cut points from the left, so a split replaces one token with
exactly two, and splits are not applied recursively. A requested split fraction
$f$ becomes $\lfloor fn \rfloor$ splits in a rollout of $n$ tokens, plus one
more with probability $fn - \lfloor fn \rfloor$, so the dose is unbiased across
rollouts; it is capped at the number of splittable tokens, and the reported
split fraction is the achieved one. The split positions are drawn at random,
or are the earliest or latest splittable positions for the location
experiments; with one split per rollout the count is exactly one.

\paragraph{The probe.} For one cell:
\begin{enumerate}
\item Sample 8 completions for each prompt from the initial checkpoint, score them,
  and normalise each prompt's rewards into advantages. Re-encode every
  completion at the requested dose and check it decodes to the same string.
\item Take canonical \texttt{token\_mean} updates on this pool, the warmup of
  Table~\ref{tab:config}, and keep the result as the checkpoint $W_0$. LoRA
  starts with one factor at zero, which leaves the other without a gradient
  until it moves.
\item Until enough informative batches are collected: draw $B = 8$ rollouts
  at random, skipping the draw if every advantage in it is zero. From $W_0$,
  with the optimiser reset, apply one update to their canonical encodings
  ($\Delta W_{\mathrm{canon}}$) and one to their re-encoded ones
  ($\Delta W_{\mathrm{re}}$). Draw $B$ further rollouts disjoint from the first,
  and apply one update to their canonical encodings ($\Delta W_{\mathrm{ref}}$).
\item Compute $c$ (Equation~\ref{eq:c}) for the pair
  $(\Delta W_{\mathrm{canon}}, \Delta W_{\mathrm{re}})$ and, with the same
  formula, for $(\Delta W_{\mathrm{canon}}, \Delta W_{\mathrm{ref}})$. The
  reported ratio $R$ (Equation~\ref{eq:R}) is the mean of the first over the
  batches where both were measured, divided by the mean of the second.
\end{enumerate}
\paragraph{Batch size.} Write the gradient-space update of a batch of $B$
rollouts as $u_b = s + n_b$ and its re-encoding displacement as
$d_b = m + e_b$, where $s$ and $m$ are shared across batches and the batch
terms $n_b$ and $e_b$ are independent with mean zero. Then $\rho =
\lVert m \rVert^2 / \mathbb{E}\lVert d_b \rVert^2$. A batch of $kB$ rollouts at
the same checkpoint has the mean of $k$ such terms, so $\mathbb{E}\lVert d
\rVert^2$ becomes $\mathbb{E}\lVert d_b \rVert^2 \bigl(\rho + (1-\rho)/k\bigr)$
while the reference's $\mathbb{E}\lVert n_b - n_{b'} \rVert^2$ is divided by
$k$. The update norms enter $c_{\mathrm{re}}$ and $c_{\mathrm{ref}}$ alike and
cancel in $R$, which gives $R(kB)^2 / R(B)^2 = 1 + (k-1)\rho$, taking the ratio
of mean norms as the ratio of root-mean-square norms.

The optimiser-free control replaces each update in step 3 by the loss gradient
at $W_0$. The cross terms of Section~\ref{sec:align} are accumulated from the
per-batch differences $\Delta W_{\mathrm{canon}} - \Delta W_{\mathrm{re}}$.

\paragraph{Completed training runs.} Rollouts are sampled once, from the
initial checkpoint, as in step 1. Both arms then start from the same initialisation and take
the same sequence of updates on the same batches in the same order, one arm on
the canonical encodings and one on the re-encoded ones, each freezing
$\pi_{\mathrm{old}}$ from its own weights at every update. Neither arm
resamples, so both train off-policy on the initial checkpoint's rollouts;
this is what makes the two arms a pair, and why these runs measure training
on fixed rollouts rather than on-policy RLVR. Final accuracy is evaluated on
held-out problems using greedy decoding.

\section{The tabular model and the counterexamples}
\label{app:simulator}
\paragraph{State.} Strings are the $2^n$ words of length $n$ over $\{a, b\}$.
A refinement shape such as $6 \to 3$ has a coarse tokenizer that cuts a word
into blocks of six characters and a fine one that cuts it into blocks of
three, so every coarse token splits into fine tokens that decode to the same
characters; the counterexamples below use $2 \to 1$. Each tokenizer has its own
tabular policy: one row of logits per token prefix, normalised by a softmax.
Both policies are built from one string distribution $p_0$ by the chain rule,
so they start from the same distribution over strings. The conditional
probability of a token given a prefix is the mass of words consistent with the
prefix and the token, divided by the mass of words consistent with the prefix.

\paragraph{The full gradient.} Take $n = 2$, a distribution $\pi$ that gives
every word positive probability, and one sampled word $ab$. A step of any of
the objectives adds $\lambda_t\,(e_{z_t} - \pi)$ to the row of each sampled token,
with $\lambda_t$ a scalar computed in any way. The coarse policy has one row over
the four words. The log-ratio of two entries of a softmax row is the
difference of their logits, so a step with scalar $\lambda$ gives
Equation~(\ref{eq:cex}) exactly. The fine policy has a root row, a row after
$a$ and a row after $b$. The step adds $\lambda_{\mathrm{root}}\,(e_a -
\pi_{\mathrm{root}})$ to the first and $\lambda_a\,(e_b - \pi_{\cdot \mid a})$ to the
second, and leaves the third, from which $\pi(ba)/\pi(bb) = \pi(a \mid
b)/\pi(b \mid b)$ is read, unchanged. If $\pi(ba) \neq \pi(bb)$, agreement
therefore requires $\lambda = 0$, which leaves the coarse distribution fixed. The
fine one is then fixed only if
$\Delta \log\bigl(\pi(a)/\pi(b)\bigr) = 2\lambda_{\mathrm{root}}\,\pi(b)$ and
$\Delta \log\bigl(\pi(b \mid a)/\pi(a \mid a)\bigr) = 2\lambda_a\,\pi(a \mid a)$
both vanish, that is, only if $\lambda_{\mathrm{root}} = \lambda_a = 0$, which proves
Proposition~\ref{prop:cex}. Nothing was assumed about how the scalars are
computed, so the argument covers pointwise weights, length normalisers,
sequence-level ratios and clipping alike. Over several inner steps each step
contributes the same expression evaluated at the current policy, so to first
order in the learning rate $\eta$ the argument applies to the sum of the scalars.

\paragraph{The symmetric exception.} If $\pi(ba) = \pi(bb)$, Equation~(\ref{eq:cex})
is zero for every $\lambda$ and the argument stops. From the uniform distribution, the
coarse step with scalar $\lambda$ gives $\pi'(ab) = e^{\lambda}/(e^{\lambda} + 3)$ and
$1/(e^{\lambda} + 3)$ to each other word. The fine step leaves
$\pi(a \mid b) = \pi(b \mid b) = \tfrac12$ and gives
$\pi'(a) = e^{\lambda_{\mathrm{root}}}/(e^{\lambda_{\mathrm{root}}} + 1)$ and
$\pi'(b \mid a) = e^{\lambda_a}/(e^{\lambda_a} + 1)$. Matching $\pi'(ba)$ requires
$\pi'(a) = (e^{\lambda} + 1)/(e^{\lambda} + 3)$, that is
$\lambda_{\mathrm{root}} = \log\bigl((e^{\lambda} + 1)/2\bigr)$; matching $\pi'(aa)$ then
requires $\pi'(b \mid a) = e^{\lambda}/(e^{\lambda} + 1)$, that is $\lambda_a = \lambda$, and
$\pi'(ab)$ follows. A rule whose steps agree for every initial distribution is
therefore zero off the set where $\pi(ba) = \pi(bb)$, which has empty interior,
and continuity in the distribution extends the zero to that set
(Corollary~\ref{cor:rule}); without continuity it need not vanish there.

\paragraph{The sampled-token update.} The simulator behind
Tables~\ref{tab:cauchy} and~\ref{tab:simulator} uses a weaker update. A batch
of (word, advantage) pairs changes each policy by adding $\eta A w_t$ to the
logit of every token of each word, in that token's own row, where $w_t$ is the
objective's weight; this keeps the sampled-token part $e_{z_t}$ of the gradient
and drops $-\pi$. As in the model experiments, a round freezes the policy and
applies several inner updates against it, which is what a weight with a ratio
is taken against. Their disagreement is the total variation between the two
updated distributions over strings. The string-space control applies
$p(s) \propto p_0(s) \exp(\eta A(s))$ and so operates directly on string
probabilities.

Under this update the coarse step moves only the logit of $ab$, so the three
unsampled words keep their relative probabilities, while the fine step changes
\[
\Delta \log \frac{\pi(aa)}{\pi(ba)} \;=\; \lambda_{\mathrm{root}} - \lambda_a\,\pi(b \mid a)
\]
to first order. A pointwise weighting sets $\lambda_{\mathrm{root}} = \eta A\,
w(\pi(a))$ and $\lambda_a = \eta A\, w(\pi(b \mid a))$. Since $\pi(a)$ and
$\pi(b \mid a)$ can be chosen independently in $(0, 1)$, agreement for every
distribution needs $w(x) = w(y)\,y$ for all $x$ and $y$: $w$ is a constant
that equals itself times every $y$, hence zero, and no continuity is needed. A
weighting that reads the rest of the sequence escapes: with coarse scalar $\lambda$,
setting $\lambda_{\mathrm{root}} = \lambda\,\pi(b \mid a)$ and $\lambda_a = \lambda$ makes the two
updated distributions agree to first order for every distribution on this
shape. Under the sampled-token update the result therefore covers pointwise
weights only.

\paragraph{The sampled word alone.} Under the sampled-token update the
first-order change of the sampled word's log-probability is
$\eta A \sum_t w_t (1 - \pi_t)$, which depends on each token's probability
alone. A coarse token $t$ and its $m$ pieces satisfy $\pi(t) = \prod_i p_i$, so an
equal change requires $w(\prod_i p_i)(1 - \prod_i p_i) = \sum_i w(p_i)(1 -
p_i)$. Writing $g(p) = w(p)(1-p)$ this is $g(\prod_i p_i) = \sum_i g(p_i)$,
Cauchy's logarithmic equation, whose continuous solutions are
$g(p) = -k \log p$: one candidate, $w(p) = -\log p / (1 - p)$, up to a common
scale, independent of the arity and inherited by nested re-encodings. It
satisfies the condition to $\IxCauchyResidual$ across arities \IxArityLo{} to
\IxArityHi{}, where \texttt{token\_sum} and \texttt{token\_mean} miss by
\IxCauchyOtherLo{} to \IxCauchyOtherHi{} (Table~\ref{tab:cauchy}). With $\pi$
uniform it changes $\log \pi(ab)$ by $\eta A \ln 4$ under both tokenizations,
but the other words by $-\tfrac13 \eta A \ln 4$ under the coarse one, against
$0$ for $aa$ and $-\eta A \ln 2$ for $ba$ and $bb$ under the fine one. A
uniform row makes $-\pi$ shift every logit equally, so these values hold under
the full gradient too.

\paragraph{Settings of Table~\ref{tab:simulator}.} Words have length $6$,
giving the shapes $6 \to 3$, $6 \to 2$, $6 \to 1$, $3 \to 1$ and $2 \to 1$. For
each of three seeds, $p_0$ is a random distribution over the $64$ words (a
uniform draw on $[0, 1]$ plus one fifth, normalised), the batch is three
distinct words with advantages $1$, $-1$ and one half, and one update with
$\eta = \IxFirstOrderLrHi{}$ and one inner step is applied; the table gives the
total variation averaged over the seeds. Repeating the $6 \to 1$ case while
halving $\eta$ five times, down to \IxFirstOrderLrLo{}, keeps the total
variation divided by $\eta$ within \IxFirstOrderDrift{}\% of
\IxFirstOrderCoefCandidate{} for the candidate and of
\IxFirstOrderCoefTokenMean{} for \texttt{token\_mean}, against
\IxFirstOrderCoefControl{} for the string-space update, as a first-order
effect should.

\section{Per-cell results}
\label{app:percell}
One record per cell is kept, named by every axis on which two cells differ. A
single loader reads them; it drops cells whose advantage density is below
\IxAdvFloor{} and cells whose two encodings did not decode to the same string,
and reports both. \IxNDropped{} cells were dropped, all \IxNDroppedAdv{} for
advantage density. It also
keeps one cell per configuration: \IxNRepeats{} cells of the learning-rate
sweep ran at the training rate itself, repeated the baseline to the last digit,
and are set aside as repeats rather than counted as further seeds.

\section{The audits of real re-encoding}
\label{app:audits}
\paragraph{Sampled IDs.} The sampled-ID audit draws \IxSidGroup{} rollouts for each of the first
\IxSidNPrompts{} prompts of a task with the probe's sampler, its settings and
its order of random draws, so that with the probe's seed its first prompts
redraw the probe's pool. Each of the \IxSidNPools{} first-seed probe pools it
covers has the advantage fraction recorded by every cell that used it, which
is evidence of a redraw rather than proof. Each completion is cut at its first
special token; its sampled IDs are compared with the encoder's encoding of
their decoded text, and the prompt's IDs followed by the completion's with the
encoding of prompt and text together; no difference crossed into the prompt.
A differing span is a merge if the
re-encoding has fewer tokens there, a split if it has more, and a moved
boundary if it has as many. Two kinds of difference are not a second
tokenization of the same bytes and are left out of
Table~\ref{tab:sampledids} in Section~\ref{sec:audit}: \IxSidUtfCut{} completions that the length limit
cut inside a multi-byte character, which decode to the replacement character
on both sides, so only a comparison of bytes detects them, and
\IxSidSpecialText{} Llama completions that spell a special token such as
\texttt{<|end\_header\_id|>} in ordinary tokens, which re-encoding parses into
the special token itself.

\paragraph{Templates and tool calls.} This audit needs only a tokenizer. Re-templating renders a conversation
with and without an appended turn and compares the prefix they share; tool-call
re-serialisation compares a call as a model emits it with the same call after
its keys are sorted or its separators compacted. A case counts as
string-preserving when the two token sequences decode to identical text.

\paragraph{Merges.} The merge audit requests merges on real \texttt{add3d} rollouts, \IxMergeCells{}
cells at requested doses of \IxMergeReqLo{} to \IxMergeReqHi{}, and reports the
edit fraction actually achieved. A merge site is an adjacent pair whose
concatenation is one vocabulary item decoding to the same text. Because BPE
applies its merges in a fixed priority order, such a pair need not be one the
encoder would join, and a moved boundary contains none, so counting merge
sites does not estimate exposure: on the resampled rollouts it flags
\IxSidMergeFlagged{}, of which \IxSidMergeHit{} differ from their re-encoding,
and misses \IxSidMergeMissed{} of the \IxSidBpeChanged{} that do.

\paragraph{Tool boundaries.} The boundary audit takes the \IxBoundarySolutions{} GSM8K test
solutions that contain a calculator annotation $\langle\langle
\mathit{expr}{=}\mathit{result}\rangle\rangle$ and treats each as a tool call:
the model writes up to and including the `=', the tool returns the result, and
the model continues from the closing brackets. The sampled sequence is the
concatenation of each segment encoded on its own, and the trainer's is the
whole solution encoded at once; both decode to the same text, and they differ
in \IxBoundaryDrifted{} of the \IxBoundarySolutions{} solutions. It runs on the
Qwen2.5 vocabulary and on the Llama-3.1 vocabulary, the latter through the
tokenizer of DeepSeek-R1-Distill-Llama-8B, which shares it. The sample strings
and calls are released with the code.

\section{Supporting tables and figures}
\label{app:heldout}

\begin{table*}[!htb]
  \centering
  \caption{The ratio $R$ to the different-batch reference at three requested
  doses: the mean over seeds, with the observed range across seeds in brackets,
  given as measured because three of the four models rest on three seeds.
  Section~\ref{sec:dose} compares the models.}
  \input{tables/scale_family}
  \label{tab:scale}
\end{table*}

\begin{table}[!htb]
  \centering
  \caption{Residual of the sampled-word condition under the sampled-token
  update (Appendix~\ref{app:simulator}), by arity. The candidate, unique up to a
  common scale, satisfies it to machine precision; the residuals for
  \texttt{token\_sum} and \texttt{token\_mean} increase with arity.}
  \adjustbox{max width=\columnwidth}{\input{tables/cauchy}}
  \label{tab:cauchy}
\end{table}

\begin{table}[!htb]
  \centering
  \caption{The weightings on the full string distribution of the tabular
  model under the sampled-token update, by refinement shape: total variation
  between the coarse and the fine updated distributions. \emph{string} is an
  update defined directly on strings, which gives the same result under both
  tokenizations; \emph{candidate} is the weighting $w(p) = -\log p/(1 - p)$
  admitted by the sampled-word condition. It yields the largest total variation distance at every
  refinement shape, by a few percent over \texttt{token\_sum} at $2 \to 1$.}
  \adjustbox{max width=\columnwidth}{\input{tables/simulator}}
  \label{tab:simulator}
\end{table}

\begin{figure}[!htb]
  \centering
  \includegraphics[width=0.6\linewidth]{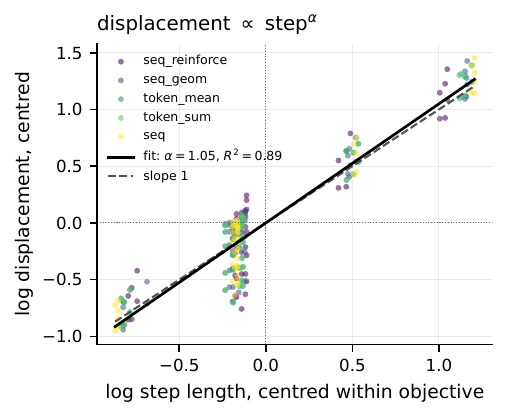}
  \caption{Log displacement against log step length, each centred within its
  objective, pooled over four probe learning rates. The black line is the
  common slope fitted through the origin, the $\alpha$ of
  Section~\ref{sec:measure}; the dashed line has slope $1$. A slope near $1$ supports normalising
  by update length over the tested range.}
  \label{fig:steps}
\end{figure}

\begin{figure*}[t]
  \centering
  \includegraphics[width=\linewidth]{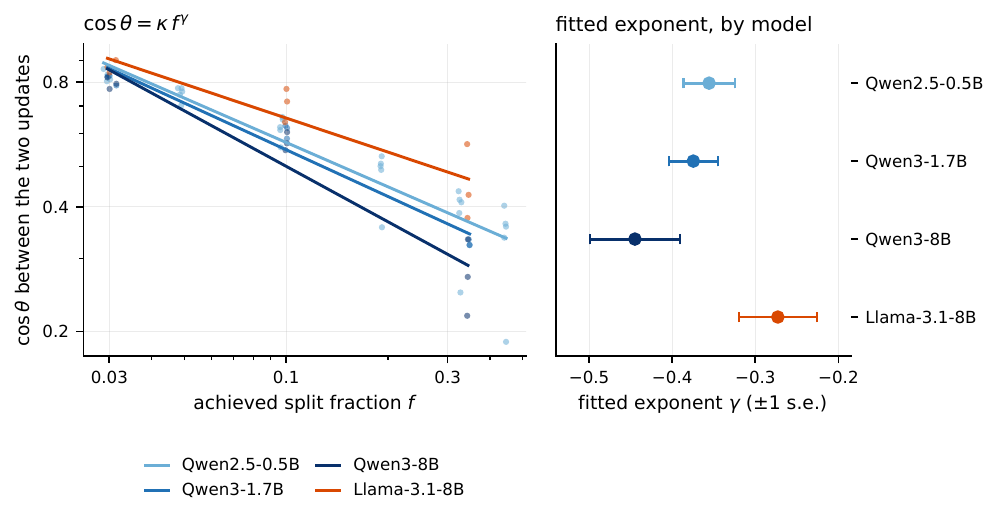}
  \caption{$\cos\theta$ between the canonical and the re-encoded update
  (\texttt{seq\_reinforce}) against the achieved split fraction, with each
  model's power-law fit (left) and its fitted exponent with $\pm1$ standard
  error (right); the two extreme exponents differ by \IxDoseExpExtremeZ{} times
  their combined standard error. The fits are an empirical description over the
  doses tested, not a mechanism, and do not extrapolate below the smallest dose
  or to zero.}
  \label{fig:doselaw}
\end{figure*}

\begin{table*}[t]
  \centering
  \caption{Completed training runs by task and dose, with accuracy on a $0$--$1$
  scale. \texttt{add3d} adds two three-digit numbers; \texttt{mul2x1} and
  \texttt{mul3x1} multiply a two- or three-digit number by a one-digit one. A
  positive gap favours the canonical arm. ``Collapsed'' counts runs whose canonical arm finished below
  \IxCollapsedBelow{} accuracy.}
  \input{tables/trained_out}
  \label{tab:trainedout}
\end{table*}

\end{document}

%% file: macros.tex
\newcommand{\IxAdvFloor}{10\%}
\newcommand{\IxAlignBatchesHi}{32}
\newcommand{\IxAlignBatchesLo}{16}

\newcommand{\IxAlignCrossFtHigh}{21}

\newcommand{\IxAlignCrossLoraHigh}{8}
\newcommand{\IxAlignCrossLoraLow}{40}

\newcommand{\IxAlignFloorMean}{0.009}
\newcommand{\IxAlignFloorSD}{0.023}
\newcommand{\IxAlignFtHigh}{0.050}
\newcommand{\IxAlignFtMid}{0.017}

\newcommand{\IxAlignFtRiseHiDose}{0.33}

\newcommand{\IxAlignFtRiseLoDose}{0.10}
\newcommand{\IxAlignFtRiseN}{5}

\newcommand{\IxAlignLoraHigh}{0.145}
\newcommand{\IxAlignLoraLow}{0.026}

\newcommand{\IxAlignLoraRiseHiDose}{0.33}

\newcommand{\IxAlignLoraRiseLoDose}{0.03}
\newcommand{\IxAlignLoraRiseN}{5}
\newcommand{\IxAlignLoraRiseP}{0.031}

\newcommand{\IxArityHi}{6}
\newcommand{\IxArityLo}{2}
\newcommand{\IxAugCanonAhead}{82}

\newcommand{\IxAugNRuns}{112}
\newcommand{\IxAugRefinedAhead}{25}
\newcommand{\IxAugTies}{5}
\newcommand{\IxBatchDoubleFront}{158}
\newcommand{\IxBatchDoubleRandom}{652}
\newcommand{\IxBoundaryChanged}{19}
\newcommand{\IxBoundaryDrifted}{13}
\newcommand{\IxBoundaryN}{8564}
\newcommand{\IxBoundarySolutions}{1301}
\newcommand{\IxBoundaryTokenFracHi}{0.0003}
\newcommand{\IxBoundaryTokenFracLo}{0.0002}

\newcommand{\IxCauchyOtherHi}{2.01}
\newcommand{\IxCauchyOtherLo}{0.25}
\newcommand{\IxCauchyResidual}{3\times 10^{-16}}
\newcommand{\IxCollapsedBelow}{0.10}
\newcommand{\IxConfAhead}{14}
\newcommand{\IxConfDose}{0.10}

\newcommand{\IxConfExDropped}{2}

\newcommand{\IxConfExGapPP}{2.9}

\newcommand{\IxConfExHiPP}{4.6}

\newcommand{\IxConfExLoPP}{1.0}

\newcommand{\IxConfExThreshold}{0.10}

\newcommand{\IxConfGapPP}{1.6}

\newcommand{\IxConfHiPP}{3.8}

\newcommand{\IxConfLoPP}{$-$1.2}
\newcommand{\IxConfN}{18}

\newcommand{\IxConfNullN}{16}

\newcommand{\IxConfNullSDPP}{9.9}

\newcommand{\IxConfSelCorr}{0.75}
\newcommand{\IxConfSignP}{0.031}
\newcommand{\IxConfSigned}{18}

\newcommand{\IxDoseExpExtremeZ}{2.4}

\newcommand{\IxDoseExpMax}{$-0.273$}
\newcommand{\IxDoseExpMin}{$-0.445$}

\newcommand{\IxDoseExpQwenHalfB}{$-0.356$}

\newcommand{\IxDoseExpSEQwenHalfB}{0.031}

\newcommand{\IxDriftTemplateN}{6}

\newcommand{\IxDriftTemplateStringKept}{2}

\newcommand{\IxDriftToolReorderN}{9}

\newcommand{\IxDriftToolReorderStringKept}{0}
\newcommand{\IxFirstOrderCoefCandidate}{0.414}
\newcommand{\IxFirstOrderCoefControl}{0.000}
\newcommand{\IxFirstOrderCoefTokenMean}{0.056}
\newcommand{\IxFirstOrderDrift}{0.4}
\newcommand{\IxFirstOrderLrHi}{0.1}
\newcommand{\IxFirstOrderLrLo}{0.003125}

\newcommand{\IxInnerEpochs}{4}

\newcommand{\IxMergeCells}{15}
\newcommand{\IxMergeEdit}{0.0005}

\newcommand{\IxMergeReqHi}{0.33}
\newcommand{\IxMergeReqLo}{0.03}

\newcommand{\IxNDosesQwenHalfB}{6}

\newcommand{\IxNDropped}{9}
\newcommand{\IxNDroppedAdv}{9}
\newcommand{\IxNNormRatio}{204}

\newcommand{\IxNRepeats}{6}
\newcommand{\IxNSameOrderNull}{17}
\newcommand{\IxNSaturated}{3}
\newcommand{\IxNSaturatedTotal}{3}
\newcommand{\IxNShapes}{5}
\newcommand{\IxNShapesInvariantWorst}{5}

\newcommand{\IxNTrainingRuns}{178}
\newcommand{\IxNormRatio}{0.983}
\newcommand{\IxNullCosLoRA}{0.308}
\newcommand{\IxNullRatioFullFT}{0.511}
\newcommand{\IxNullRatioGSMEIGHTK}{0.655}

\newcommand{\IxNullRatioLoRA}{0.764}
\newcommand{\IxNullRatioLoRAAtGSMDoses}{0.717}

\newcommand{\IxNullRatioQwenHalfBHi}{0.945}
\newcommand{\IxNullRatioQwenHalfBLo}{0.475}

\newcommand{\IxNullValueLoRA}{1.160}

\newcommand{\IxOneAlignBack}{0.026}

\newcommand{\IxOneAlignFront}{0.160}
\newcommand{\IxOneAlignFrontCross}{7}

\newcommand{\IxOneAlignFrontN}{5}
\newcommand{\IxOneAlignFrontRatio}{14.8}
\newcommand{\IxOneAlignRandom}{0.037}

\newcommand{\IxOneTokenFrontBackRatio}{1.65}

\newcommand{\IxOneTokenOneIn}{25}
\newcommand{\IxOneTokenRandom}{0.556}
\newcommand{\IxOneTokenSplit}{0.0407}

\newcommand{\IxPosDiffHi}{0.111}
\newcommand{\IxPosDiffLo}{0.245}
\newcommand{\IxPosDiffMid}{0.264}
\newcommand{\IxPosDiffN}{5}

\newcommand{\IxPosDoseEquiv}{0.092}
\newcommand{\IxPosDoseFold}{2.3}

\newcommand{\IxPosFirstLo}{0.616}

\newcommand{\IxPosLastLo}{0.371}

\newcommand{\IxPosRandomLo}{0.475}

\newcommand{\IxProbeBatch}{8}

\newcommand{\IxSidBlocks}{274}
\newcommand{\IxSidBlocksMerge}{201}
\newcommand{\IxSidBlocksSame}{65}
\newcommand{\IxSidBlocksSplit}{8}
\newcommand{\IxSidBpeChanged}{219}
\newcommand{\IxSidEditsPerRollout}{2.6}
\newcommand{\IxSidGroup}{8}
\newcommand{\IxSidMaxTokenFrac}{0.0008}
\newcommand{\IxSidMaxTokenPct}{0.08\%}
\newcommand{\IxSidMergeFlagged}{1135}
\newcommand{\IxSidMergeHit}{168}
\newcommand{\IxSidMergeMissed}{51}
\newcommand{\IxSidNModels}{4}
\newcommand{\IxSidNPools}{5}
\newcommand{\IxSidNPrompts}{256}
\newcommand{\IxSidNRollouts}{2048}

\newcommand{\IxSidPoolAddChanged}{1}
\newcommand{\IxSidPoolAddN}{1536}
\newcommand{\IxSidPoolGsmChanged}{25}
\newcommand{\IxSidPoolGsmN}{768}
\newcommand{\IxSidRestPct}{0.39\%}
\newcommand{\IxSidSmallestDose}{0.03}
\newcommand{\IxSidSpecialText}{3}
\newcommand{\IxSidTopPctHi}{5.2\%}
\newcommand{\IxSidTopPctLo}{4.1\%}
\newcommand{\IxSidUtfCut}{3}
\newcommand{\IxSpaceDoseFoldGradient}{2.8}
\newcommand{\IxSpaceDoseFoldUpdate}{2.0}

\newcommand{\IxSpaceDoseHiGradient}{0.972}

\newcommand{\IxSpaceDoseHiUpdate}{0.945}

\newcommand{\IxSpaceDoseLoGradient}{0.344}

\newcommand{\IxSpaceDoseLoUpdate}{0.475}

\newcommand{\IxSpaceFrontBackGradient}{2.29}
\newcommand{\IxSpaceFrontBackUpdate}{1.65}
\newcommand{\IxSpaceN}{30}

\newcommand{\IxSpaceRTwo}{0.983}
\newcommand{\IxSpaceSlope}{1.31}
\newcommand{\IxSplitCeilBare}{0.521}
\newcommand{\IxSplitCeilChat}{0.044}
\newcommand{\IxStepAlpha}{1.049}

\newcommand{\IxStepRatioLoRA}{1.329}

\newcommand{\IxWarmupAlignHi}{0.092}
\newcommand{\IxWarmupAlignLo}{0.145}
\newcommand{\IxWarmupAlignMid}{0.096}

\newcommand{\IxWarmupDropNEightThirtyTwo}{5}

\newcommand{\IxWarmupHi}{64}
\newcommand{\IxWarmupLo}{8}
\newcommand{\IxWarmupMid}{32}
\newcommand{\IxWarmupRatioHi}{16.1}
\newcommand{\IxWarmupRatioLo}{11.5}
\newcommand{\IxWarmupSeeds}{5}

%% file: tables/space.tex
\begin{tabular}{lrrrrc}
  \toprule
  condition & update & gradient & shift & ratio & lower \\
  \midrule
  dose 0.03 & 0.475 & 0.344 & -0.131 & 0.72 & 5/5 \\
  dose 0.10 & 0.730 & 0.658 & -0.072 & 0.90 & 5/5 \\
  dose 0.33 & 0.945 & 0.972 & +0.027 & 1.03 & 1/5 \\
  one token, first & 0.715 & 0.614 & -0.102 & 0.86 & 5/5 \\
  one token, random & 0.556 & 0.426 & -0.129 & 0.77 & 5/5 \\
  one token, last & 0.433 & 0.268 & -0.165 & 0.62 & 5/5 \\
  \bottomrule
\end{tabular}

%% file: tables/objective.tex
\begin{tabular}{lrr}
  \toprule
  objective & LoRA & full fine-tuning \\
  \midrule
  \texttt{seq\_reinforce} & 0.883 & 0.697 \\
  \texttt{seq\_geom} & 0.950 & 0.767 \\
  \texttt{token\_mean} & 0.956 & 0.870 \\
  \texttt{token\_sum} & 0.958 & 0.870 \\
  \texttt{seq} & 0.972 & 0.708 \\
  \midrule
  step ratio & 1.329 & 1.045 \\
  cells & 30 & 30 \\
  \bottomrule
\end{tabular}

%% file: tables/alignment.tex
\begin{tabular}{llrrr}
  \toprule
  & dose & re-encoding & control & seeds \\
  \midrule
  LoRA & 0.03 & 0.026 & 0.011 & 5 \\
   & 0.10 & 0.067 & 0.011 & 5 \\
   & 0.33 & 0.145 & 0.011 & 5 \\
  full fine-tuning & 0.10 & 0.017 & 0.007 & 5 \\
   & 0.33 & 0.050 & 0.007 & 5 \\
  \midrule
  one token, first & 0.04 & 0.160 & 0.011 & 5 \\
  one token, random & 0.04 & 0.037 & 0.011 & 5 \\
  one token, last & 0.04 & 0.026 & 0.011 & 5 \\
  \bottomrule
\end{tabular}

%% file: tables/sampled_ids.tex
\begin{tabular}{lrrr}
  \toprule
  model & addition & GSM8K & GSM8K, chat \\
  \midrule
  Qwen2.5-0.5B & 4 (0.20) & 83 (4.05) & 106 (5.18) \\
  Qwen3-1.7B & 1 (0.05) & 8 (0.39) & 3 (0.15) \\
  Qwen3-8B & 0 (0.00) & 1 (0.05) & 0 (0.00) \\
  Llama-3.1-8B & 7 (0.34) & 5 (0.24) & 1 (0.05) \\
  \bottomrule
\end{tabular}

%% file: tables/config.tex
\begin{tabular}{lp{0.62\linewidth}}
  \toprule
  setting & value \\
  \midrule
  \multicolumn{2}{l}{\emph{policy}} \\
  \quad models & Qwen/Qwen2.5-0.5B-Instruct, Qwen/Qwen3-1.7B, Qwen/Qwen3-8B, meta-llama/Llama-3.1-8B-Instruct \\
  \quad adapter & LoRA, rank 16, $\alpha = 2r$, on the attention projections \\
  \quad full fine-tuning & every parameter, base widened to float32 \\
  \multicolumn{2}{l}{\emph{optimisation}} \\
  \quad optimiser & AdamW, state reset before each update, per arm and batch \\
  \quad learning rate & $5 \times 10^{-5}$ \\
  \quad probe learning rate & equal to the learning rate unless swept \\
  \quad clip $\epsilon$ & 0.2 \\
  \quad inner epochs & 4 \\
  \multicolumn{2}{l}{\emph{sampling}} \\
  \quad rollouts per prompt & 8 \\
  \quad temperature, top-$p$ & 1.0, 0.95 \\
  \quad max new tokens & 24 on the synthetic tasks, 200 on GSM8K \\
  \quad prompts per cell & 32, 48, 96 \\
  \quad evaluation problems & 200 \\
  \multicolumn{2}{l}{\emph{measurement}} \\
  \quad canonical warmup rounds & 8, 32, 64 \\
  \quad batches per cell & 16, 32, 64 \\
  \quad seeds & 0, 1, 2, 3, 4 \\
  \bottomrule
\end{tabular}

%% file: tables/config_confirm.tex
\begin{tabular}{lp{0.62\linewidth}}
  \toprule
  setting & value \\
  \midrule
  checkpoint & \texttt{Qwen/Qwen2.5-0.5B-Instruct}, no chat template \\
  adapter & LoRA, rank 16 \\
  training problems & 96 from the GSM8K training split, per seed \\
  evaluation problems & 200 from the GSM8K test split, per seed \\
  rollout pool & 8 per problem, 768 in all, sampled once \\
  sampling & temperature 1.0, top-$p$ 0.95 \\
  max new tokens & 200, for sampling and evaluation \\
  evaluation & greedy decoding \\
  objective & \texttt{token\_mean} \\
  updates & 300, each on 8 rollouts drawn from the pool \\
  inner epochs per update & 1 \\
  learning rate & $5 \times 10^{-5}$ \\
  split fraction & requested 0.10, achieved 0.100 \\
  seeds & 18 pairs, seeds 0--17 \\
  \bottomrule
\end{tabular}

%% file: tables/scale_family.tex
\begin{tabular}{lrrccc r}
  \toprule
  & & & \multicolumn{3}{c}{requested split fraction} & \\ \cmidrule(lr){4-6}
  model & params & vocab & 0.03 & 0.10 & 0.35 & seeds \\
  \midrule
  Qwen2.5-0.5B & 0.5B & 151.9k & 0.47 \scriptsize[0.42, 0.54] & 0.73 \scriptsize[0.69, 0.84] & 0.95 \scriptsize[0.90, 1.10] & 5 \\
  Qwen3-1.7B & 1.7B & 151.9k & 0.56 \scriptsize[0.54, 0.57] & 0.82 \scriptsize[0.80, 0.86] & 1.09 \scriptsize[1.03, 1.15] & 3 \\
  Qwen3-8B & 8B & 151.9k & 0.52 \scriptsize[0.46, 0.56] & 0.78 \scriptsize[0.71, 0.81] & 1.03 \scriptsize[0.94, 1.08] & 3 \\
  \bfseries Llama-3.1-8B & 8B & 128k & 0.39 \scriptsize[0.33, 0.43] & 0.61 \scriptsize[0.54, 0.67] & 0.84 \scriptsize[0.75, 0.90] & 3 \\
  \bottomrule
\end{tabular}

%% file: tables/cauchy.tex
\begin{tabular}{rrrr}
  \toprule
  arity & \texttt{token\_sum} & \texttt{token\_mean} & candidate \\
  \midrule
  2 & 0.250 & 0.252 & $8 \times 10^{-17}$ \\
  3 & 0.623 & 0.377 & $1 \times 10^{-16}$ \\
  4 & 1.055 & 0.440 & $2 \times 10^{-16}$ \\
  5 & 1.520 & 0.471 & $2 \times 10^{-16}$ \\
  6 & 2.007 & 0.486 & $3 \times 10^{-16}$ \\
  \bottomrule
\end{tabular}

%% file: tables/simulator.tex
\begin{tabular}{lrrrr}
  \toprule
  shape & string & \texttt{token\_sum} & \texttt{token\_mean} & candidate \\
  \midrule
  $6 \to 3$ & 0.000 & 0.0114 & 0.0054 & \bfseries 0.0264 \\
  $6 \to 2$ & 0.000 & 0.0179 & 0.0056 & \bfseries 0.0327 \\
  $6 \to 1$ & 0.000 & 0.0317 & 0.0056 & \bfseries 0.0413 \\
  $3 \to 1$ & 0.000 & 0.0238 & 0.0047 & \bfseries 0.0288 \\
  $2 \to 1$ & 0.000 & 0.0173 & 0.0031 & \bfseries 0.0179 \\
  \bottomrule
\end{tabular}

%% file: tables/trained_out.tex
\begin{tabular}{llrrrrr}
  \toprule
  task & split fraction & $n$ & mean gap & median & canonical ahead & collapsed \\
  \midrule
  \texttt{add3d} & $\geq 0.20$ & 16 & +0.315 & +0.082 & 15/16 & 0 \\
  \texttt{gsm8k} & $0.05$--$0.20$ & 17 & +0.017 & +0.040 & 12/17 & 1 \\
  \texttt{gsm8k} & $\geq 0.20$ & 11 & +0.044 & +0.075 & 8/11 & 1 \\
  \texttt{mul2x1} & $0.05$--$0.20$ & 3 & +0.025 & +0.025 & 3/3 & 0 \\
  \texttt{mul2x1} & $\geq 0.20$ & 5 & +0.040 & +0.045 & 5/5 & 0 \\
  \texttt{mul3x1} & $<0.05$ & 30 & +0.008 & +0.007 & 16/30 & 0 \\
  \texttt{mul3x1} & $0.05$--$0.20$ & 9 & +0.051 & +0.015 & 8/9 & 0 \\
  \texttt{mul3x1} & $\geq 0.20$ & 21 & +0.189 & +0.050 & 15/21 & 1 \\
  \bottomrule
\end{tabular}